\documentclass[lettersize,journal]{IEEEtran}
\usepackage{amsmath,amsfonts}
\usepackage{algorithmic}
\usepackage{algorithm}
\usepackage{array}
\usepackage[caption=false,font=normalsize,labelfont=sf,textfont=sf]{subfig}
\usepackage{textcomp}
\usepackage{stfloats}
\usepackage{url}
\usepackage{verbatim}
\usepackage{graphicx}
\usepackage{cite}
\usepackage{bm}
\usepackage{booktabs}
\usepackage{multirow}
\usepackage{colortbl}
\usepackage{threeparttable}
\usepackage{hyperref}
\usepackage{siunitx}

\definecolor{color_1st}{rgb}{1.0, 0.7, 0.7}
\definecolor{color_2nd}{rgb}{0.945, 0.87, 0.74}
\definecolor{color_3rd}{rgb}{1, 0.98, 0.84}

\begin{document}

\title{Duplex-GS: Proxy-Guided Weighted Blending for Real-Time Order-Independent Gaussian Splatting}

\author{Weihang Liu,
        Yuke Li,
        Yuxuan Li,
        Jingyi Yu,~\IEEEmembership{Fellow,~IEEE,}
        Xin Lou,~\IEEEmembership{Senior Member,~IEEE,}
\thanks{This work has been submitted to the IEEE for possible publication. Copyright may be transferred without notice, after which this version may no longer be accessible.}
}

\markboth{Journal of \LaTeX\ Class Files,~Vol.~14, No.~8, August~2021}%
{Shell \MakeLowercase{\textit{et al.}}: A Sample Article Using IEEEtran.cls for IEEE Journals}

\IEEEpubid{0000--0000/00\$00.00~\copyright~2021 IEEE}

\maketitle

\begin{abstract}
3D Gaussian Splatting (3DGS) achieves photorealistic rendering but incurs substantial overhead due to the global sorting required for $\alpha$-blending, which results in noticeable ``popping" artifacts and hinders deployment on edge devices. While sort-free Order-Independent Transparency (OIT) methods circumvent sorting, they introduce ``transparency" artifacts and suffer from inefficiency due to the absence of physical constraints.
To address these limitations, we present Duplex-GS, a dual-hierarchy framework that leverages proxy-guided spatial organization and a novel hybrid renderer that combines $\alpha$-blending with reformulated Weighted-Sum Rendering (WSR). We introduce explicit ellipsoidal cell proxies to encapsulate local Gaussians, which enables efficient proxy-level rasterization. This strategy drastically reduces the overhead associated with global sorting. 
Furthermore, we propose a physically grounded WSR scheme with cell-level early termination, which retrieves the physical constraints absent in prior OIT-based 3DGS methods, effectively eliminating both popping and transparency artifacts.
Extensive experiments on diverse real-world datasets demonstrate the robustness of Duplex-GS across scenarios spanning multi-scale training views and large-scale environments. Quantitatively, our method delivers high-fidelity real-time rendering, outperforming prior OIT-based 3DGS methods by $1.5\times$\textendash\,$4\times$ in speed, while reducing radix-sort cost by $29.8\%\,$\textendash\,$86.9\%$ compared with conventional $\alpha$-blending without compromising visual quality.

\end{abstract}


\section{Introduction}
\IEEEPARstart{N}{ovel} view synthesis (NVS) and real-time rendering have attracted significant attention in both academia and industry, with broad applications in video games, virtual reality (VR)~\cite{VRNeRF}, and 3D scene reconstruction~\cite{CityGo, tcsvt_3_FRPGS, yan2024street, 2DGS, tcsvt_25_1, tcsvt_25_4, tcsvt_25_3, tcsvt_4}. Inspired by the groundbreaking Neural Radiance Field (NeRF)~\cite{NeRF}, learning-based methods have substantially advanced the realism of digital content, narrowing the gap between the virtual and real worlds. However, these methods often entail considerable computational overhead. The emergence of 3D Gaussian Splatting (3DGS)~\cite{vanilla_3DGS, tcsvt_survey} marks another milestone, enabling photorealistic and real-time rendering of radiance fields via rasterization, a paradigm akin to the graphics pipeline in modern graphics processing units (GPUs).

\begin{figure*}
    \centering
    \includegraphics[width=1.0\linewidth]{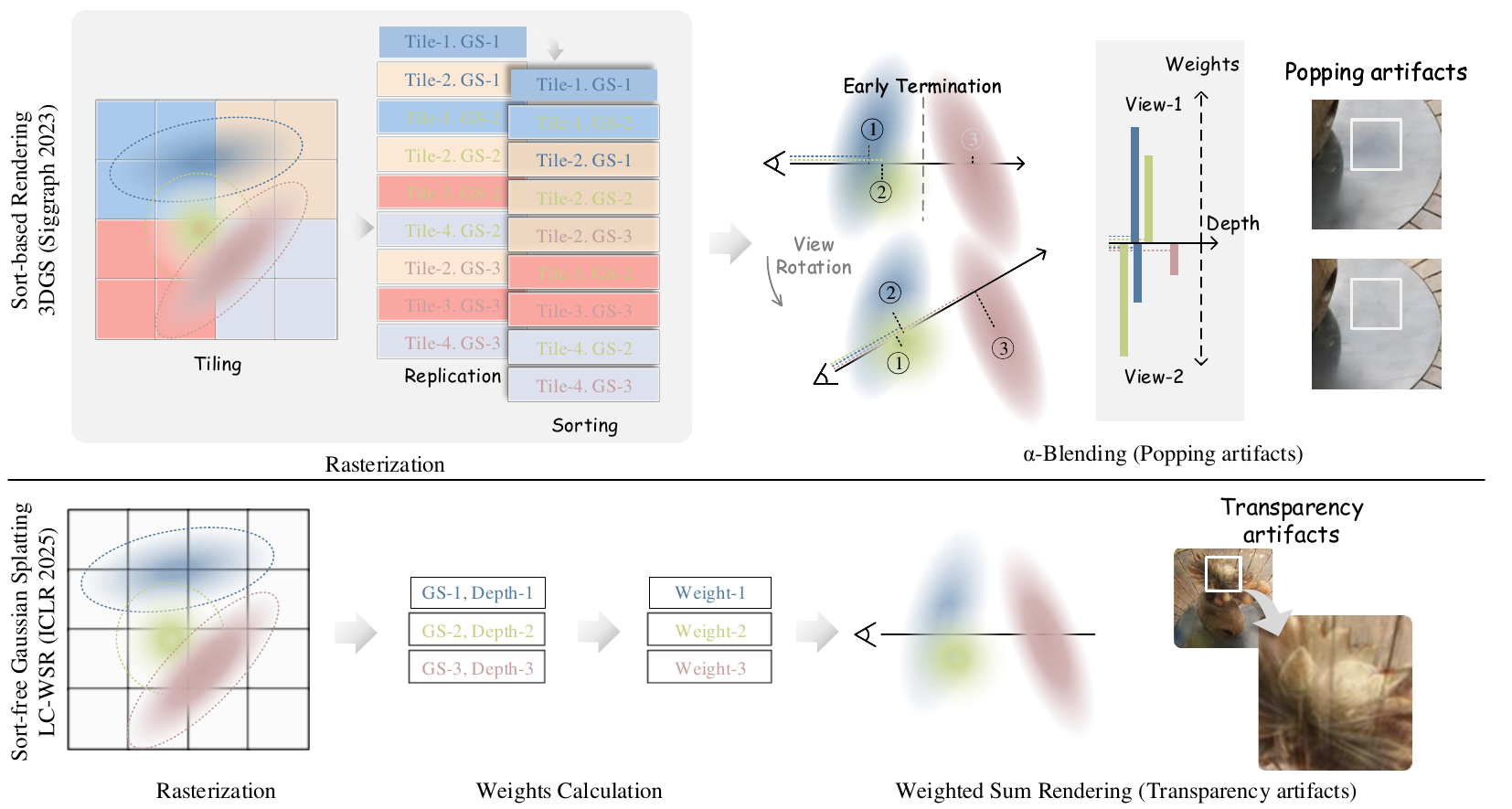}
    \caption{\textbf{Motivation of Duplex-GS.} 
    (Top) 3DGSs are rendered via $\alpha$-blending for a long tile-Gaussian sequence. However, sudden changes in viewing angle can cause abrupt changes in the blending order and the corresponding weights, resulting in noticeable popping artifacts. 
    (Bottom) The sort-free LC-WSR method eliminates the sorting stage through WSR paradigm. However, the absence of physical constraints leads to transparency artifacts and efficiency degradation, as blending continues unnecessarily for opaque surfaces.}
    \label{fig:motivation}
\end{figure*}

Despite the impressive reconstruction and rendering capabilities of 3DGS, a primary limitation is its need for an ever-increasing number of Gaussian primitives to fit all training views. This results in substantial and irregular memory access during rendering, as costly view-dependent sorting operations over long sequences impose considerable computational and memory overhead~\cite{ISSCC2025_GS_Tsinghua, ISSCC2025_GS_KAIST}. Moreover, the sorting process can introduce visible ``popping" artifacts~\cite{stopthepop}, as illustrated in Fig.~\ref{fig:motivation}. Collectively, these limitations hinder the scalability of 3DGS for large-scale applications and restrict its deployment on resource-constrained platforms such as mobile phones and VR devices.

More recently, sort-free Gaussian Splatting~\cite{Sort-Free} has incorporated Order-Independent Transparency (OIT)~\cite{OIT1, OIT2} into the Gaussian Splatting framework, eliminating the need for sorting in the rasterization pipeline by employing Weighted Sum Rendering (WSR)~\cite{WSR_OIT}. This effectively eliminates popping artifacts and has been successfully demonstrated on mobile devices~\cite{Sort-Free}. However, as illustrated in Fig.~\ref{fig:motivation}, the absence of physical constraints leads to ``transparency" artifacts and increased rendering time, particularly in large-scale scenes, due to the accumulation of redundant Gaussian primitives and lack of early termination when encountering opaque objects in the blending phase.

\begin{figure*}[t]
    \centering
    \includegraphics[width=1.0\linewidth]{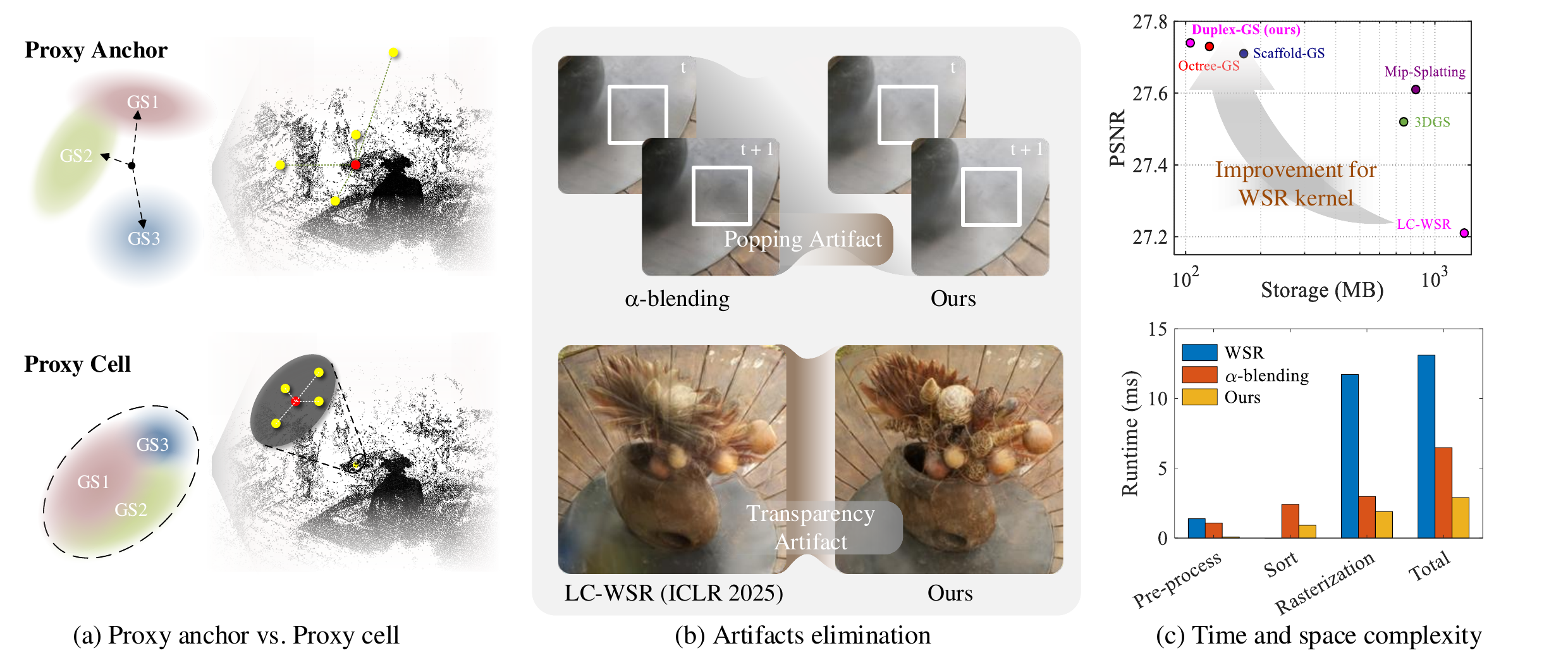}
    \caption{
    \textbf{Overview of the proposed Duplex-GS.} Our approach is built on a dual-hierarchy structure that introduces ellipsoidal cells as proxies for local Gaussians.
    (a) \textbf{Explicit Cell Proxies.} Unlike abstract anchor points, our geometrically defined proxy cells are directly utilized for rasterization, guiding the subsequent Gaussian blending process.
    (b) \textbf{Hybrid Rendering Paradigm.} By synergistically combining $\alpha$-blending with physically grounded WSR, our method simultaneously eliminates both popping and transparency artifacts.
    (c) The integration of cell rasterization (Sec.~\ref{sec:layerI}) and physically grounded WSR (Sec.~\ref{sec:layerII}) achieves significant improvements over prior OIT-based Gaussian splatting methods in both rendering accuracy and computational efficiency.
    }
    \label{fig:intro}
\end{figure*}

Neural Gaussian frameworks~\cite{Scaffold-GS, Octree-GS} enhance storage efficiency by using ``anchor" points to encode local Gaussians via shared MLPs. However, they still require sorting the decoded Gaussians for $\alpha$-blending, which is costly on resource-constrained devices~\cite{ISSCC2025_GS_KAIST}. Moreover, the lack of explicit spatial constraints allows Gaussians to deviate from their anchors (Fig.~\ref{fig:intro}), complicating the representation.

\IEEEpubidadjcol
This paper presents Duplex-GS, a novel framework for modeling 3D scenes using explicit proxies with defined visible regions and a tailored hybrid rendering paradigm. 
Our approach integrates the benefits of $\alpha$-blending and WSR to eliminate artifacts while maintaining real-time performance.
First, we encode scenes using ellipsoidal proxies, termed ``cells", which replace the anchors used in the original neural Gaussian framework. By constraining neural Gaussians to reside within their respective cells, rasterization is performed at the cell level to guide subsequent Gaussian blending. Second, activated cells are decoded into Gaussians via shared MLPs. By restoring physical constraints absent in previous OIT-based 3DGS~\cite{Sort-Free}, we introduce a physically grounded WSR strategy that enables early termination and replaces non-commutative blending operations with a parallelizable weighted sum.

This hybrid design offers two key advantages. 
First, the significantly smaller population of cell proxies compared to Gaussians enables our cell-level rasterization strategy to drastically reduce the memory and computational overhead of global radix sorting.
Second, we restore physical constraints absent in prior OIT-based 3DGS method, which facilitates cell-level early termination and removes both popping and transparency artifacts simultaneously.

In summary, the contributions of this work are as follows:
\begin{itemize}
\item{
We propose a dual-layered Gaussian hierarchy comprising ellipsoidal proxy cells and 3D Gaussians. This structure enables artifact-free, real-time 3D reconstruction while maintaining the parallelizability of WSR.
}
\item{\textbf{Cell rasterization (layer I)}: 
We introduce ellipsoidal cells with adaptive geometric properties to spatially organize neural Gaussians. A novel cell-level rasterization strategy is proposed, which significantly reduces the memory and computational overhead associated with global radix sorting.
}
\item{\textbf{Physically grounded WSR blending (layer II)}: 
We incorporate physical constraints into the WSR method to blend local Gaussians, inherently avoiding popping and transparency artifacts. Building on this, we introduce a cell-level early termination scheme that further accelerates rendering.
}
\item{
We provide a high-performance CUDA implementation of our WSR-based Gaussian splatting framework. Experimental results demonstrate that our method achieves competitive rendering speed and superior visual fidelity on consumer-grade hardware.
}
\end{itemize}

\section{Related Work}
\subsection{Neural Representation for 3D Scene}
Traditional scene reconstruction methods like SfM~\cite{SfM1, SfM2} and MVS~\cite{MVS1, MVS2} struggle with texture-less regions and lighting variations. NeRF~\cite{NeRF} enabled photorealistic reconstruction from sparse images, with subsequent efficient variants~\cite{NGP, CoARF, TensoRF} making real-time rendering feasible on edge devices~\cite{mobileNeRF, TGRS_R4}.
3DGS~\cite{vanilla_3DGS} marked a breakthrough by using explicit anisotropic Gaussians for fast rendering, though its computational and memory demands limit its deployment on resource-constrained platforms. A clear trend shows that more explicit representations improve speed at the cost of memory, leading to hybrid approaches~\cite{NGP, Plenoxel, tcsvt_1_mesh_aligned, Scaffold-GS, BGT, HybridGS, HybridGS_icml} that balance this trade-off. For example, Instant-NGP~\cite{NGP} uses a multi-resolution hash table to reduce MLP burden, Scaffold-GS~\cite{Scaffold-GS} encodes neural Gaussians via anchors to reduce model size while maintaining quality.
However, these methods still rely on the conventional $\alpha$-blending pipeline that requires sorting Gaussians by depth. This step complicates implementation and introduces visual popping artifacts \cite{stopthepop}.

\subsection{Efficient Neural Rendering}
While NeRF is primarily constrained by high computational demands, the practical deployment of 3DGS is bottlenecked by its extensive number of Gaussian primitives and frequent memory access requirements~\cite{ISSCC2025_GS_KAIST, ISSCC2025_GS_Tsinghua}. 
Existing efforts to optimize the original $\alpha$-blending pipeline in 3DGS broadly fall into three categories: 1) Applying model compression techniques, such as Level-of-Detail (LOD) strategy~\cite{Octree-GS, HierarchicalGS, letsgo, CityGS}, encoding schemes~\cite{Scaffold-GS, Compact_3DGS, SOGS}, as well as pruning and quantization~\cite{Compressed_3DGS, Light_3DGS, CoQ}. 
2) Optimizing critical operators through CUDA-based implementations~\cite{huang2025seele, gssplat, DISTWAR, FlashGS, DashGaussian} or dedicated hardware acceleration units~\cite{GScore}.
3) Replacing Gaussian kernels with more efficient representation such as linear kernals~\cite{3DLS} and half-Gaussian kernel~\cite{3D-HGS}.

A few works have attempted to revise the core rendering paradigm itself by exploring efficient alternatives to $\alpha$-blending. 
StochasticSplats~\cite{kv2025stochasticsplats} introduces a stochastic transparency paradigm which enables efficient rendering using OpenGL shaders at the risk of controllable quality degradation.
Sort-free Gaussian Splatting~\cite{Sort-Free} introduces a Weighted Sum Rendering (WSR) pipeline, which eliminates popping artifacts and achieves faster rendering on mobile devices for small-scale scenes. Nevertheless, its lack of physical constraints leads to transparency artifacts and introduces unnecessary computation when rendering opaque surfaces.

\subsection{Order Independent Transparency}
OIT~\cite{OIT1, OIT2} is a foundational graphics technique integrated into modern GPUs to enable the correct compositing of transparent fragments without requiring explicit depth sorting prior to rasterization. Early approaches, such as Depth Peeling~\cite{Dual_Depth_peeling}, A-buffer~\cite{A_buffer}, and Multi-Layer Alpha Blending (MLAB)~\cite{MLAB}, successfully achieve OIT but often incur significant time or memory overhead. Weighted Blended OIT~\cite{WSR_OIT} offers a more efficient alternative by approximating the integration of transparency in a single render pass, storing weighted sums of color and opacity with carefully designed blending weights. And this method has been applied to 3DGS rendering~\cite{Sort-Free}.
However, the absence of physical constraints in\cite{Sort-Free} degrades both visual quality and rendering efficiency. In contrast, the proposed Duplex-GS framework restores the physically grounded transmittance in proxy-level, enabling early termination to halt compositing upon encountering fully opaque surfaces with weighted blended OIT.
As a result, Duplex-GS attains rendering quality and speed on par with state-of-the-art (SOTA) 3DGS models on high-end GPUs, while exhibiting strong potential for deployment on resource-constrained edge devices by alleviating the huge overhead of the view-adaptive sorting stage.

\begin{figure*}[t]
    \centering
    \includegraphics[width=1.0\linewidth]{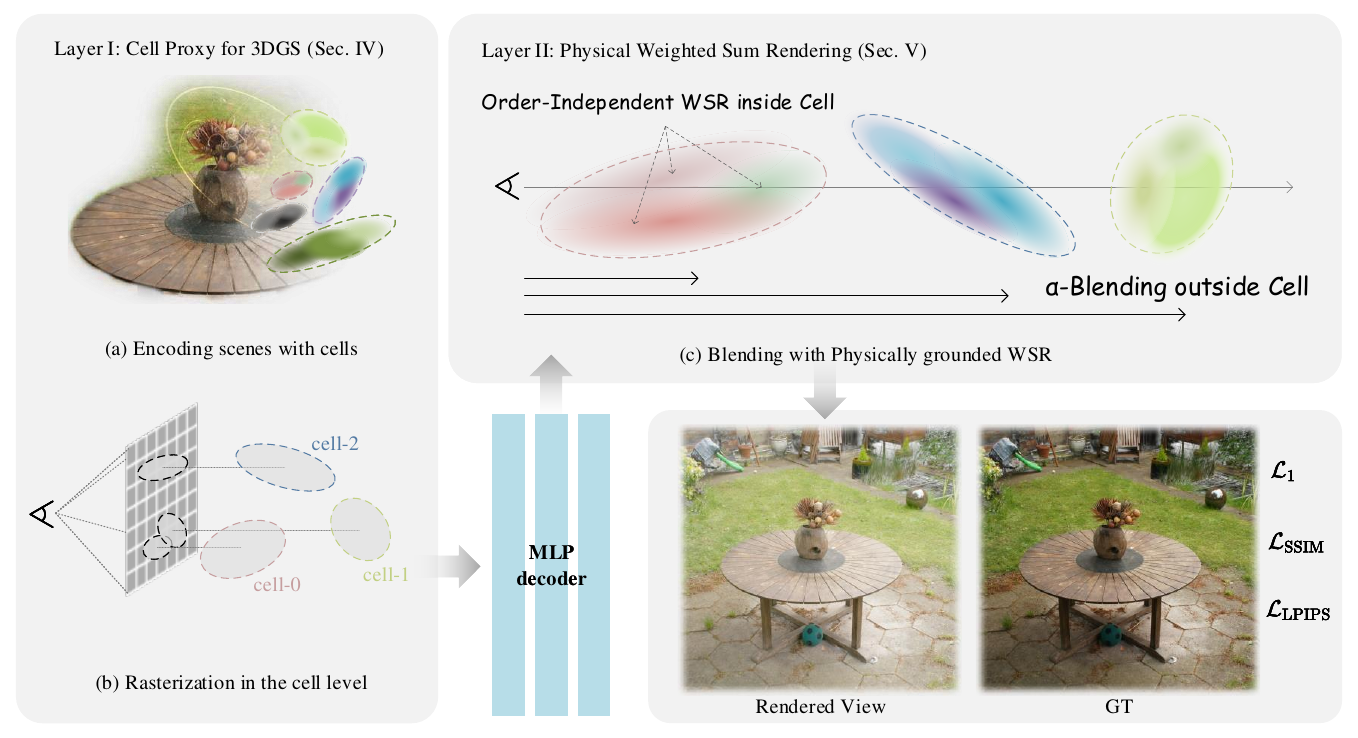}
    \caption{\textbf{Illustration of the proposed Duplex-GS pipeline.} 
    (Layer I) Ellipsoidal cells act as proxies to spatially organize local Gaussians. This structure enables efficient cell-level rasterization, significantly reducing the overhead of view-adaptive global sorting.
    (Layer II) Intersected cells are decoded into Gaussians and blended with the proposed physically grounded WSR. The final color is computed by first determining cell-level weights via $\alpha$-blending, then aggregating the Gaussians inside each cell in an order-independent WSR manner.
    }
    \label{fig:method}
\end{figure*}

\section{Preliminaries and Overview}

\subsection{Preliminaries}
\subsubsection{Neural Gaussian}
3DGS~\cite{vanilla_3DGS} models scenes using anisotropic 3D Gaussian primitives and renders images through a rasterization based paradigm. Building upon 3DGS, 
Scaffold-GS \cite{Scaffold-GS} introduces anchors as local encoding proxies, which emit neural Gaussians decoded from latent features to effectively represent local scene structure. Specifically, each anchor emits a predefined set of $K$ neural gaussians, whose centers are given by:
\begin{equation}
    \label{eq:scaffold_offset}
    \bm{\mu}_{n,k} = \bm{x}_{n} + \mathcal{O}_{n,k}^{\text{pos}} \cdot \bm{S}_n,
\end{equation}
where $n\in\{1,2,...,N\}$ indexes the anchors, $k\in\{1,2,...K\}$ indexes the neural Gaussians per anchor, $\bm{x}_n$ denotes the center of anchor-$n$, and $\bm{\mu}_{n,k}$
represents the center of the emitted neural Gaussian-$k$.
$\mathcal{O}_{n,k}^{\text{pos}}$ is the learned offset scaled by $\bm{S}_n$.
Additionally, the opacities of the $K$ neural Gaussians are decoded from the latent feature $\bm{f}_n$, viewing distance $\delta_{n}$ and direction $\bm{r}_{n}$ via a scene-wise MLP
\begin{equation}
    \{\sigma_{n,1},...,\sigma_{n,K}\} = F_\sigma(\bm{f}_n, \delta_{n}, \bm{r}_{n}),
\end{equation}
where $F_\sigma$ is the shared opacity decoder where the output is activated by $\text{Tanh}$.
Similarly, the other attributes of each neural Gaussians, including color $\bm{c}_{n,k}$, scales $\bm{s}_{n,k}$ and quaternions $\bm{q}_{n,k}$, are 
predicted by separate MLPs $F_c$, $F_s$, $F_q$, respectively.
Consequently, rendering proceeds via a tile based Gaussian rasterization technique, which employs radix sort 
to blend $M$ ordered points
\begin{equation}
    \label{eq:alpha_blending}
    \bm{C}(\bm{x}^{\prime}) = \sum_{i\in M}T_i^\text{$\alpha$-blending}\bm{c}_i\alpha_i, \quad \alpha_i=\sigma_iG^{\prime}_i(\bm{x}^{\prime}),
\end{equation}
where $\bm{x}^{\prime}$ denotes the 2D coordinates of the queried pixel, $G^\prime_i(\bm{x}^{\prime})$ is the projected 2D gaussian kernel, and $T_i^\text{$\alpha$-blending}$ denotes the transmittance
\begin{equation}
    \label{eq:transmittance}
    T_i^\text{$\alpha$-blending} = \prod_{j=1}^{i-1} (1-\alpha_j).
\end{equation}

\subsubsection{Weighted Sum Rendering}
Inspired by OIT, a widely adopted technique for rendering non-opaque media in traditional graphics pipelines, sort-free Gaussian Splatting integrates this concept into 3DGS, producing the final image as
\begin{equation}
    \label{eq:WSR}
    \bm{C} = \frac{\bm{c}_Bw_B+\sum_{i=1}^N\bm{c}_i\alpha_iw(d_i)}
             {w_B+\sum_{i=1}^N\alpha_iw(d_i)},
\end{equation}
where $\bm{c}_B$ and $w_B$ denote the background color and the learnable background weights respectively, and $d_i$ indicates the depths of the $i$-th 3D Gaussians.

By employing this WSR formulation, the non-commutative, computationally intensive sorting stage, traditionally required for compositing, is eliminated. This significantly enhances the potential for parallelization and accelerates rendering.
Specifically, the Linear Correction Weighted-Sum Rendering (LC-WSR) function $w(d_i)$ achieve best performance which is defined as
\begin{equation}
    w(d_i) = T_i^\text{LC-WSR}\cdot v_i,
\end{equation}
\begin{equation}
    \label{eq:LC-WSR}
    T_i^\text{LC-WSR} = \max \left (0, 1-\frac{d_i}{\tau} \right ),
\end{equation}
where $\tau$ and $v_i$ are the learnable parameters. This function formulates the decay trend w.r.t. the depths of the rendering Gaussians.

\subsection{Overview}
We introduce a novel framework featuring a tailored encoding and rendering paradigm, as illustrated in Fig. \ref{fig:method}. Our approach is structured in two primary stages.
First, 3D scenes are modeled using ellipsoidal cells, which act as proxies that emit Gaussians with coherent features inside their coverage.
This is coupled with a hybrid rendering paradigm that integrates the advantages of $\alpha$-blending and WSR, establishing a \textit{globally ordered yet locally unordered} scene structure. Specifically, rasterization is first performed at the proxy layer to compute a blending weight for each cell in $\alpha$-blending manner. The emitted Gaussians are then rendered using a revised WSR paradigm.

This pipeline conducts sorting at the cell level, establishing a coarse spatial order for patches of Gaussians. This strategy drastically reduces the computational burden of performing a radix sort on every individual Gaussian primitive (Sec.~\ref{sec:layerI}). Within each cell, Gaussians remain locally unsorted, making the order-independent WSR paradigm naturally suitable for their blending. This design eliminates popping artifacts, as the rendering result is locally invariant to the Gaussian order. Furthermore, by reintroducing physical constraints into the OIT-based renderer, a physically grounded transmittance decay is guaranteed, as shown in Fig.~\ref{fig:T}, which removes the transparency artifacts inherent in the vanilla OIT-based Gaussian Splatting~\cite{Sort-Free} (Sec. \ref{sec:layerII}) as well as enables early termination in cell-level to accelerate rendering.


\section{Layer I: Cell Proxy for 3DGS}
\label{sec:layerI}
While neural Gaussian frameworks~\cite{Scaffold-GS, Octree-GS} have achieved remarkable reconstruction accuracy, they still rely on conventional $\alpha$-blending operations as the introduced proxy anchors are points without concrete geometric interpretation.
We introduce ellipsoidal cells with explicit visible regions as the proxies. Based on this representation, a cell rasterization strategy is introduced, enabling efficient and lightweight rasterization.

\begin{figure}
    \centering
    \includegraphics[width=1.0\linewidth]{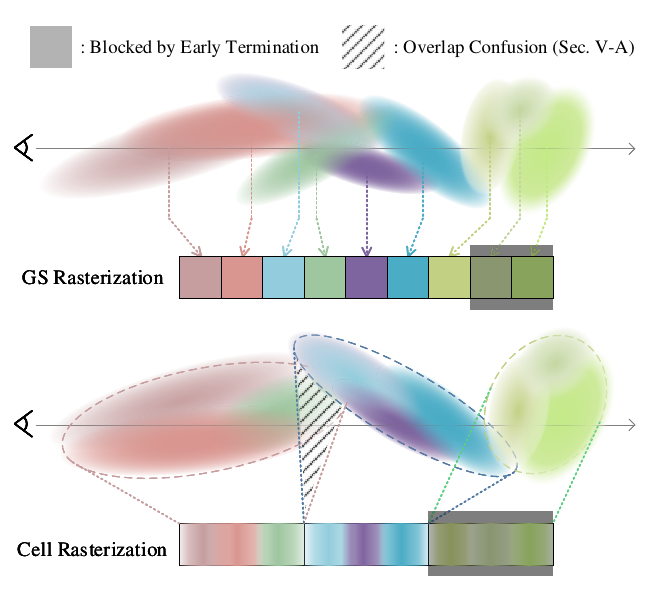}
    \caption{Comparison between rasterization of GS and cell. Unlike the original $\alpha$-blending which processes each Gaussian individually, our method performs rasterization and early termination at the coarser cell level, where each cell encapsulates multiple Gaussians. The order confusion caused by overlap among cells is discussed in Sec.~\ref{sec:V-A}.}
    \label{fig:cell_search}
\end{figure}

\subsection{Cell Proxy}
Unlike anchors, which are points lacking geometric structure, each cell is represented as an ellipsoid that covers several neural 3D Gaussians whose physical properties are dynamically decoded from a trainable feature.

Initialization begins with SfM points $\mathbf{P}\in\mathbb{R}^{N\times3}$. We generate the initial cells from these points, centering each one in accordance with the method described in \cite{Octree-GS}.
Every cell is parameterized by position $\bm{x}_{n}$, scales $\bm{S}_n$, quaternions $\bm{Q}_n$, defining its visible region, and a calibration scalar $v_n$ (discussed in Sec.~\ref{sec:V-A}). In the dual-layered hierarchy, each cell is further associated with a latent feature vector $\bm{f}_n$ that encodes local scene information, and emits $K$ neural Gaussians as described in Eq.~\ref{eq:scaffold_offset}.
The attributes of these $K$ neural Gaussians, including color $\bm{c}_{n,k}$, scales ratio $\mathcal{O}_{n,k}^{\text{scale}}$ and quaternions $\bm{q}_{n,k}$, are decoded from the latent feature $\bm{f}_n$, along with the viewing distance $\delta_{n}$ and viewing direction $\bm{d}_{n}$, via shared scene-wise MLPs.

\begin{equation}
    \{\bm{c}^{n,1}, ..., \bm{c}^{n,K}\} = F_c(f_n, \delta_{n}, d_{n}),
\end{equation}
\begin{equation}
    \{\mathcal{O}^{\text{scale}}_{n,1}, ..., \mathcal{O}^{\text{scale}}_{n,K}\} = F_s(f_n, \delta_{n}, d_{n}),
\end{equation}
\begin{equation}
    \{\bm{q}_{n,1}, ..., \bm{q}_{n,K}\} = F_q(f_n, \delta_{n}, d_{n}).
\end{equation}
To ensure that each Gaussian center lies within the visible region of its corresponding cell, we constrain the offset parameter $\mathcal{O}_{n,k}^{\text{pos}}$ to the range $[-1,1]$. 
We propose that cells are likely to emit Gaussians with similar geometric properties. So scales of Gaussians are derived under the supervision of $\bm{S}_n$:
\begin{equation}
    \label{eq:nerual_scales}
    \bm{s}_{n,k} = 
    \mathcal{O}_{n,k}^{\text{scale}} * \bm{S}_n, 
\end{equation}
where the output of $F_s$ is activated by sigmoid to restrict the ratio $\mathcal{O}_{n,k}^\text{scale}$ to the range $(0, 1)$.
Based on Eq.\ref{eq:nerual_scales}, the scales of cells can be directly optimized. While the quaternions unable to optimize in the similar way, as they denote the orientation of the clustered Gaussians. To align regions of cell and Gaussians, the cell quaternions are updated as the weighted average of the associated Gaussians' orientation
\begin{equation}
    \label{eq:quaternion}
    \bm{Q}_n = \text{norm}(\sum_k^K \bm{q}_{n,k} * \sigma_{n,k}).
\end{equation}
Additionally, to restrict that the Gaussian regions fully locate inside the cell region, the maximum of $\mathcal{O}_{n,k}^{\text{scale}}$ is further clamped by
\begin{equation}
\label{eq:scale_res}
    \mathcal{O}_{n,k}^{\text{scale}} = \min\left(\mathcal{O}_{n,k}^{\text{scale}},
    1 - ||\mathcal{O}_{n,k}^{\text{pos}}||
    \right).
\end{equation}
As illustrated in Fig.~\ref{fig:method}, the proposed neural Gaussians are constrained to reside strictly within their corresponding cell regions. Leveraging this property, the cell projections can effectively guide the Gaussian blending process.

\subsection{Cell Rasterization}
Conventional $\alpha$-blending renders views by rasterizing 3D Gaussians onto the image plane via splatting, a process that necessitates a global Gaussian traversal and sort. We fundamentally reformulate this pipeline by introducing a cell-level rasterization technique, where ellipsoidal cells, rather than individual Gaussians, serve as the primary rendering primitives. As illustrated in Fig.~\ref{fig:cell_search}, this approach significantly reduces the number of primitives handled. Specifically, each cell defines a bounded visible region; only cells intersecting the viewing frustum are selected and decoded into neural 3D Gaussians.

This proxy-guided design inherently ensures spatial alignment between the cell distribution and the underlying Gaussians, confining computationally intensive rasterization operations exclusively to the coarser cell level. Operating within a tile-based paradigm, visible cells are assigned to corresponding tiles, and a radix sort is applied to the cell sequence. Since the number of cells is orders of magnitude smaller than that of Gaussians, this strategy yields substantial reductions in both memory consumption and computational overhead for the rasterization process.

\begin{figure}
    \centering
    \includegraphics[width=1.0\linewidth]{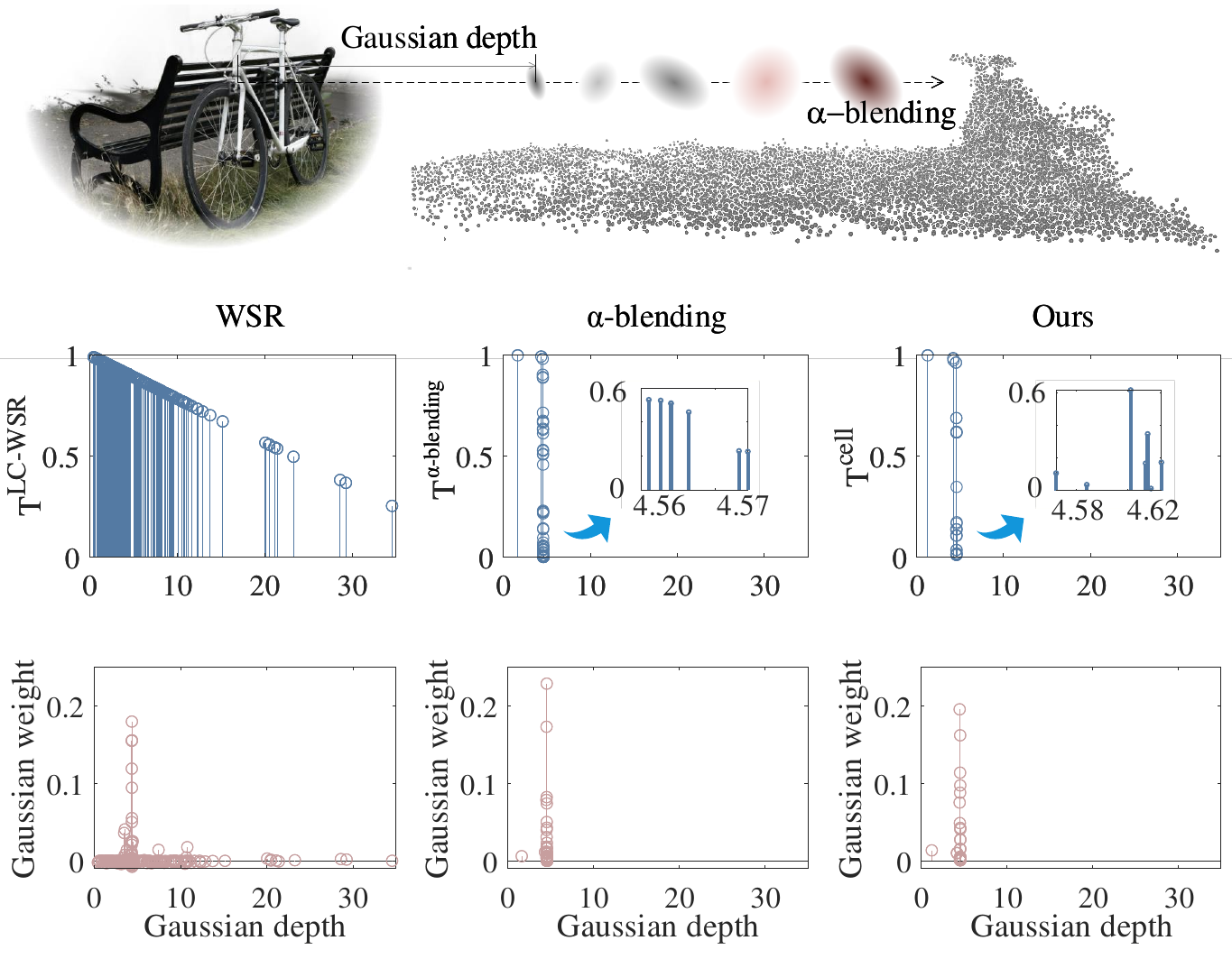}
    \caption{Comparison of three different transmittance: LC-WSR (Eq.~\ref{eq:LC-WSR}), vanilla $\alpha$-blending (Eq.~\ref{eq:transmittance}) and ours (Eq.~\ref{eq:dulpex_T}). 
    By reintroducing physical constraints, our model produces a transmittance curve similar to that of $\alpha$-blending, establishing it as a reliable criterion for early termination. Furthermore, the proposed transmittance is not strictly monotonic with depth, confirming the local order-independent nature of Gaussian blending in our framework.}
    \label{fig:T}
\end{figure}

\section{Layer II: Physical Weighted Sum Rendering}
\label{sec:layerII}
Duplex-GS reconstructs 3D scenes using ellipsoidal cells, which enable rasterization and sorting at a coarse cell level to significantly reduce rasterization overhead. However, this approach forgoes the explicit per-Gaussian depth sorting required by the original $\alpha$-blending pipeline, which is grounded in physical light transport models. This introduces an ambiguity in the blending order. The Sort-free Gaussian method~\cite{Sort-Free}, an OIT extension of 3DGS, demonstrates that high-fidelity view synthesis can be achieved through a weighted sum of Gaussians, governed by mathematical rather than physical constraints. This paradigm is highly suitable for graphics rendering pipelines in modern GPU and resource-constrained devices. Nevertheless, abandoning the physical blending model introduces transparency artifacts that degrade visual quality, and the lack of early termination further compromises rendering efficiency.

In this section, we integrate the OIT based rendering paradigm with a cell based Gaussian model, retrieving physical constraints for WSR based 3DGS.

\subsection{Physical OIT Blending for 3DGS}
\label{sec:V-A}

Although the WSR paradigm~\cite{Sort-Free} provides a mathematically-grounded rendering model, it lacks a physical foundation. Consequently, the blending weight $\alpha_i$ in Eq.~\ref{eq:WSR} does not correspond to a physically meaningful opacity and is unconstrained, often exceeding unity. This formulation not only introduces transparency artifacts but also precludes the use of early termination when blending opaque surfaces, limiting its efficiency.

Within this coarse-ordered structure obtained from cell rasterization, we redefine $\alpha$ as an opacity weighted by the Gaussian kernel (Eq.~\ref{eq:alpha_blending}), constraining its value to the physically plausible range of $[0,1]$. The final rendered view $\bm{C}$ is computed according to the weighted sum rendering formula:
\begin{equation}
    \label{eq:hybrid_renderer}
    \bm{C} = \frac{\sum_{n=1}^{N}w_n\sum_{k=1}^K\alpha_{n,k}\bm{c}_{n,k}}
    {\sum_{n=1}^Nw_n\sum_{k=1}^K\alpha_{n,k}},
\end{equation}
where $w_n$ denotes the weight of cell-$n$, derived from the cell-level sorting results.
A sophisticated WSR kernel for each cell, $w_n$ can then be defined as:
\begin{equation}
    w_n = v_n \cdot T_n^\text{cell},
\end{equation}
where $T_n^\text{cell}$ denotes the cell-level transmittance, which is efficiently obtained by sorting the sparsely distributed cells, and it decays based on the accumulated opacity from preceding cells:
\begin{equation}
\label{eq:dulpex_T}
    T_n^\text{cell} = \prod_{i=1}^{n-1} \prod_{k=1}^K (1-v_i \cdot \alpha_{i,k}),
\end{equation}
where a learnable calibration scalar $v_i$ is introduced to 
address the ambiguity arising from non-uniform spatial distribution and potential inter-cell overlaps, as illustrated in Fig.~\ref{fig:cell_search}. 
Since cell occlusion relationships are view-dependent, we parameterize $v_i$ using spherical harmonic coefficients to capture its directional variation.
The resulting Eq.~\ref{eq:hybrid_renderer} shows that our framework produces model which is ordered outside cells and order-independent inside each cell.

The opacity distribution in a 3D scene is typically sparse, implying that the accumulated transmittance is physically modeled by a step-function which suddenly changes encountering points with high opacity rather than a gradual decay. To validate this theory, we provide three transmittance paradigms and the resulting Gaussians weights in Fig.~\ref{fig:T}. 
LC-WSR simulates transmittance with a linear function of Gaussian depth. However, this model is physically inconsistent, as it is derived from purely mathematical constraints. In contrast, our method produces a weight distribution similar to vanilla $\alpha$-blending without requiring a precise per-Gaussian depth order, which eliminates transparency artifacts for that the transmittance decays rapidly when encountered opaque object. Furthermore, the rendering of Gaussian primitives is locally order-independent, as the transmittance does not strictly decrease with depth. This property forms the cornerstone for the elimination of popping artifacts. 

\subsection{Early Termination for Cell Based 3DGS}
As defined in Eq.~\ref{eq:dulpex_T}, the cell-level transmittance $T_n^\text{cell}$ decays towards zero during front-to-back rendering, eliminating transparency artifacts. This provides a natural mechanism for early termination: the blending process is halted in cell level once $T_n^\text{cell}$ falls below a predefined threshold $\epsilon$, as illustrated in Fig.~\ref{fig:cell_search}. 
This approach effectively accelerates rendering.
Unlike prior WSR methods for 3DGS, our formulation restores explicit physical meaning to the opacity parameter $\sigma_{n,k}$ by constraining it to the physically plausible range of $[0,1]$, which means the empirical threshold can be directly transferred to our model. This enforcement of physical plausibility subsequently facilitates more effective pruning and regularization during the training process.

\section{Optimization Strategies for Cell Based Gaussian Splatting}
\label{sec:alignment}
The cell search rasterization inevitably introduces redundant 3D Gaussians during the blending stage, due to the discrepancies that arise between the spatial intervals of cells and the emitted neural Gaussians as shown in Fig.~\ref{fig:false_positive_skip}a. As the efficiency of cell search rasterization is contingent upon the degree of coherence between the visible regions of a cell and its associated Gaussians, we provide several engineering-driven strategies to facilitate training of Duplex-GS.

\subsection{False Positive Skip}
\begin{figure}
    \centering
    \includegraphics[width=1.0\linewidth]{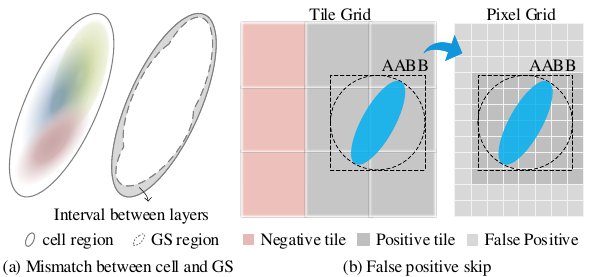}
    \caption{Illustration of mismatch between cell and Gaussian. (a) The interval between layers influence the efficiency of cell rasterization. (b) False positive regions are skipped to alleviate the speed degradation from the mismatch.}
    \label{fig:false_positive_skip}
\end{figure}
Tile-based Rendering (TBR) is widely adopted in modern GPUs and is also utilized in 3DGS rasterization. However, this approach often introduces a substantial number of redundant false-positive Gaussians. This problem is further exacerbated in cell-based Gaussian frameworks due to the spatial mismatch between cell regions and the actual coverage of Gaussians. To mitigate the associated performance degradation, we introduce a pixel-level radius based culling strategy to eliminate unnecessary Gaussian kernel computations. Specifically, as illustrated in Fig.~\ref{fig:false_positive_skip}, a lightweight axis-aligned bounding box (AABB) check is employed: Gaussians are excluded from the blending operation if the pixel lies outside their respective AABBs. This efficient early-exit mechanism significantly reduces redundant computation and accelerates the overall rendering process.

\subsection{Geometric Correction via Neural Gaussians}
In addition to skipping false-positive Gaussians to improve efficiency, maximizing the overlap between each cell and its emitted neural Gaussians further reduces the number of redundant Gaussians. To achieve better geometric alignment, we dynamically define the cell’s quaternion as a weighted average of the quaternions associated with its emitted Gaussians, as described in Eq.~\ref{eq:quaternion}. Moreover, to ensure an even spatial distribution of Gaussians within each cell, the cell centers are periodically updated using
\begin{equation}
    \bm{x}_{n} = \frac{\sum_{k=1}^{K}\bm{\mu}_{n,k}}{K}.
\end{equation}
As cell scales are differentiable via Eq.~\ref{eq:nerual_scales}, the three-dimensional covariances and positions of first-layer cells are learned in accordance with the updates to the second-layer Gaussians.

\subsection{Cell Position Reset}
Following adaptive density control~\cite{vanilla_3DGS}, we grow new cells at Gaussians with high view-space positional gradients. Unlike abstract anchors~\cite{Scaffold-GS, Octree-GS}, our cells have explicit geometry. 
To decouple existing cells from newly introduced ones, the offsets of the significant Gaussians are reset to zero after cell growth, i.e., their positions are reinitialized to the centers of the corresponding cells.


\subsection{Perceptual Loss}
The loss function for optimizing 3DGS is typically formulated as a combination of $\ell_1$ and SSIM~\cite{SSIM} losses, which respectively capture pixel-level reconstruction error and structural similarity. 
As proven in GAN-based models, these pixel-wise image differences results in solutions with high signal-to-noise ratios, which are perceptually rather smooth and less convincing~\cite{Perceptual_loss1, Perceptual_loss2}.
We additionally incorporate a perceptual loss term for training Gaussian based models. The overall loss is defined as
\begin{equation}
    \label{eq:loss}
    \mathcal{L} = (1-\lambda_1-\lambda_2) \mathcal{L}_1 + \lambda_1 \mathcal{L}_{\text{SSIM}} + \lambda_2\mathcal{L}_{\text{LPIPS}},
\end{equation}
where $\mathcal{L}_1$, $\mathcal{L}_{\text{SSIM}}$ and $\mathcal{L}_{\text{LPIPS}}$ denote $\ell_1$, structural similarity and perceptual loss respectively.
While the inclusion of the perceptual loss yields rendering results with improved perceptual quality, its computation relies on convolutional neural networks (CNNs), which increases the training time. To address this, we compute the perceptual loss every 20 training iterations, which substantially alleviates the additional computational overhead without noticeably increasing the overall training time.

\IEEEpubidadjcol

\section{Experimental Results}
\begin{table*}[tb]
    \setlength{\tabcolsep}{3pt}
    \caption{
    Quantitative comparisons on real-world dataset \cite{mipnerf360, tandt, db}. Duplex-GS achieves competitive rendering quality while reducing model size compared to the baselines. The best performance for each metric is highlighted. 
    }
    \label{tab:res1}
    \begin{center}
        \begin{threeparttable}
        \begin{tabular*}{\linewidth}{@{\extracolsep{\fill}}p{3.4cm}cccc cccc cccc@{}}
            \toprule
            \multirow{2}{*}{\centering Method}
            & \multicolumn{4}{c}{Mip-NeRF360}
            & \multicolumn{4}{c}{Tanks\&Temples}
            & \multicolumn{4}{c}{DeepBlending}
            \\
            \cmidrule[0.5pt](lr){2-5} \cmidrule[0.5pt](lr){6-9} \cmidrule[0.5pt](lr){10-13} 
            {\ }
            & $\text{PSNR}_\uparrow $
            & $\text{SSIM}_\uparrow $
            & $\text{LPIPS}_\downarrow $
            & $\text{Storage}_\downarrow $
            & $\text{PSNR}_\uparrow $
            & $\text{SSIM}_\uparrow $
            & $\text{LPIPS}_\downarrow $
            & $\text{Storage}_\downarrow $
            & $\text{PSNR}_\uparrow $
            & $\text{SSIM}_\uparrow $
            & $\text{LPIPS}_\downarrow $
            & $\text{Storage}_\downarrow $
            \\
            \midrule
            Mip-NeRF360$^1$ \cite{mipnerf360}
            & 27.69
            & 0.792
            & 0.237
            & -
            & 22.22
            & 0.759
            & 0.257
            & -
            & 29.40
            & 0.901
            & 0.245
            & -
            \\
            3DGS$^1$ \cite{vanilla_3DGS}
            & 27.52
            & 0.813
            & 0.222
            & 750.2\,MB
            & 23.57
            & 0.845
            & 0.180
            & 431.4\,MB
            & 29.61
            & 0.900
            & 0.251
            & 662.7\,MB
            \\
            Mip-Splatting$^1$ \cite{mip_splatting}
            & 27.61
            & \cellcolor{color_1st}0.816
            & \cellcolor{color_1st}0.215
            & 838.4\,MB
            & 23.96
            & 0.856
            & 0.171
            & 500.4\,MB
            & 29.56
            & 0.901
            & 0.243
            & 736.8\,MB
            \\
            \midrule
            Scaffold-GS$^1\,(K=10)$ \cite{Scaffold-GS}
            & 27.73
            & 0.812
            & 0.226
            & 171.0\,MB
            & 24.09
            & 0.858
            & 0.165
            & 147.7\,MB
            & 30.42
            & \cellcolor{color_1st}0.912
            & 0.246
            & 111.2\,MB
            \\
            Scaffold-GS$^1\,(K=5)$ \cite{Scaffold-GS}
            & 27.74
            & 0.811
            & 0.230
            & 205.3\,MB
            & 24.53
            & 0.863
            & 0.162
            & 177.7\,MB
            & 30.26
            & 0.911
            & 0.242
            & 143.4\,MB
            \\
            Octree-GS$^1\,(K=10)$ \cite{Octree-GS}
            & 27.88
            & \cellcolor{color_1st}0.816
            & 0.216
            & 142.7\,MB
            & \cellcolor{color_1st}24.60
            & 0.864
            & 0.157
            & 77.8\,MB
            & \cellcolor{color_1st}30.44
            & 0.911
            & 0.239
            & 95.5\,MB
            \\
            Octree-GS$^1\,(K=5)$ \cite{Octree-GS}
            & 27.73
            & 0.813
            & 0.227
            & 124.5\,MB
            & 24.47
            & 0.861
            & 0.168
            & \cellcolor{color_1st}73.9 MB
            & 30.06
            & 0.908
            & 0.250
            & \cellcolor{color_1st}75.1 MB
            \\
            \midrule
            LC-WSR$^2$ \cite{Sort-Free}
            & 27.21
            & 0.800
            & 0.219
            & 1312.0\,MB
            & 23.22
            & 0.832
            & 0.186
            & 672.7\,MB
            & 29.90
            & 0.901
            & 0.243
            & 810.7\,MB
            \\
            StocSplats$^3$(16~SPP) \cite{kv2025stochasticsplats}
            & 26.25
            & 0.714
            & 0.351
            & -
            & -
            & -
            & -
            & -
            & -
            & -
            & -
            & -
            \\
            \midrule
            Ours $(K=10)$
            & \cellcolor{color_1st}27.90
            & 0.813
            & 0.216
            & 152.2\,MB
            & 24.26
            & \cellcolor{color_1st}0.867
            & 0.150
            & 124.8\,MB
            & 30.30
            & 0.910
            & 0.249
            & 89.7\,MB
            \\
            Ours $(K=5)$
            & 27.74
            & 0.802
            & 0.218
            & \cellcolor{color_1st}104.4\,MB
            & 24.32
            & \cellcolor{color_1st}0.867
            & \cellcolor{color_1st}0.138
            & 112.2\,MB
            & 30.26
            & 0.909
            & \cellcolor{color_1st}0.236
            & 94.7\,MB
            \\
            \bottomrule
        \end{tabular*}
        \begin{tablenotes}
            \footnotesize
            \item[1] Experiments are conducted using the official public repository, with no modifications except for iteration settings as described in Sec.~\ref{sec:implementation_details}.
            \item[2] Experiments are conducted with our own implementation, which has been recognized by the author of \cite{Sort-Free}. Codes are available at \href{https://github.com/LiYukeee/sort-free-gs}{\url{https://github.com/LiYukeee/sort-free-gs}}.
            \item[3] Results from original paper (code unavailable). Missing data reflects source omissions.
        \end{tablenotes}
        \end{threeparttable}
    \end{center}
\end{table*}

\begin{table*}[tb]
    \setlength{\tabcolsep}{3pt}
    \caption{
    Quantitative comparisons on BungeeNeRF~\cite{bungeenerf} and VR-NeRF~\cite{VRNeRF} datasets. Additional experiments assess performance on multi-scale rendering and the reconstruction of intricate indoor details. Duplex-GS achieves high-quality rendering with compact model sizes. The best result for each metric is highlighted.
    }
    \label{tab:res2}
    \begin{center}
        \begin{threeparttable}
        \begin{tabular*}{\linewidth}{@{\extracolsep{\fill}}p{3.4cm}cccc cccc cccc@{}}
            \toprule
            \multirow{2}{*}{\centering Method}
            & \multicolumn{4}{c}{BungeeNeRF}
            & \multicolumn{4}{c}{VR-NeRF: apartment}
            & \multicolumn{4}{c}{VR-NeRF: kitchen}
            \\
            \cmidrule[0.5pt](lr){2-5} \cmidrule[0.5pt](lr){6-9} \cmidrule[0.5pt](lr){10-13}
            {\ }
            & $\text{PSNR}_\uparrow $
            & $\text{SSIM}_\uparrow $
            & $\text{LPIPS}_\downarrow $
            & $\text{Storage}_\downarrow $
            & $\text{PSNR}_\uparrow $
            & $\text{SSIM}_\uparrow $
            & $\text{LPIPS}_\downarrow $
            & $\text{Storage}_\downarrow $
            & $\text{PSNR}_\uparrow $
            & $\text{SSIM}_\uparrow $
            & $\text{LPIPS}_\downarrow $
            & $\text{Storage}_\downarrow $
            \\
            \midrule
            3DGS$^1$ \cite{vanilla_3DGS}
            & 27.71
            & 0.915
            & 0.099
            & 1654.1\,MB
            & 30.98
            & 0.922
            & 0.212
            & 368.1\,MB
            & 31.73
            & 0.933
            & 0.185
            & 443.6\,MB
            \\
            Mip-Splatting$^1$ \cite{mip_splatting}
            & 28.21
            & \cellcolor{color_1st}0.922
            & 0.096
            & 1108.8\,MB
            & 31.49
            & 0.929
            & 0.201
            & 442.8\,MB
            & 32.19
            & 0.939
            & 0.177
            & 484.3\,MB
            \\
            \midrule
            Scaffold-GS$^1\,(K=10)$ \cite{Scaffold-GS}
            & 27.36
            & 0.901
            & 0.122
            & 328.5\,MB
            & 30.70
            & 0.927
            & 0.182
            & 840.1\,MB
            & 31.20
            & 0.932
            & 0.166
            & 658.5\,MB
            \\
            Scaffold-GS$^1\,(K=5)$ \cite{Scaffold-GS}
            & 27.32
            & 0.898
            & 0.130
            & 353.7\,MB
            & 31.30
            & 0.932
            & 0.170
            & 740.4\,MB
            & 31.05
            & 0.930
            & 0.171
            & 593.2\,MB
            \\
            Octree-GS$^1\,(K=10)$ \cite{Octree-GS}
            & 27.07
            & 0.901
            & 0.117
            & 351.0\,MB
            & 31.19
            & 0.922
            & 0.212
            & 217.7\,MB
            & 32.33
            & 0.939
            & \cellcolor{color_1st}0.157
            & 372.7\,MB
            \\
            Octree-GS$^1\,(K=5)$ \cite{Octree-GS}
            & 26.93
            & 0.894
            & 0.133
            & 329.9\,MB
            & 31.34
            & 0.921
            & 0.224
            & \cellcolor{color_1st}165.3 MB
            & 31.40
            & 0.929
            & 0.181
            & 403.0\,MB
            \\
            \midrule
            LC-WSR$^2$ \cite{Sort-Free}
            & 26.94
            & 0.903
            & 0.111
            & 1670.2\,MB
            
            & 31.84
            & \cellcolor{color_1st}0.935
            & \cellcolor{color_1st}0.169
            & 1543.2\,MB
            
            & 32.32
            & 0.937
            & 0.158
            & 1482.7 MB
            
            \\
            \midrule
            Ours $(K=10)$
            & \cellcolor{color_1st}28.80
            & 0.920
            & \cellcolor{color_1st}0.093
            & 319.2\,MB
            & 31.79
            & 0.927
            & 0.218
            & 191.7\,MB
            & 32.17
            & 0.934
            & 0.191
            & \cellcolor{color_1st}176.2\,MB
            \\
            Ours $(K=5)$
            & 28.66
            & 0.914
            & 0.098
            & \cellcolor{color_1st}262.8\,MB
            & \cellcolor{color_1st}32.27
            & 0.932
            & 0.178
            & 299.0\,MB
            & \cellcolor{color_1st}32.38
            & \cellcolor{color_1st}0.940
            & \cellcolor{color_1st}0.157
            & 269.8\,MB
            \\
            \bottomrule
        \end{tabular*}
        \begin{tablenotes}
            \footnotesize
            \item[1] Experiments are conducted using the official public repository, with no modifications except for iteration settings as described in Sec.~\ref{sec:implementation_details}.
            \item[2] Experiments are conducted with our own implementation, which has been recognized by the author of \cite{Sort-Free}. Codes are available at \href{https://github.com/LiYukeee/sort-free-gs}{\url{https://github.com/LiYukeee/sort-free-gs}}.
        \end{tablenotes}
        \end{threeparttable}
    \end{center}
\end{table*}

\begin{table}[tb]
    \setlength{\tabcolsep}{3pt}
    \caption{
    Quantitative comparisons on MatrixCity. Duplex-GS achieves highest accuracy across all metric in such large-scale aerial scenarios while maintaining competitive efficiency.
    }
    \label{tab:MatrixCity}
    \begin{center}
        \begin{threeparttable}
        \begin{tabular*}{\linewidth}{@{\extracolsep{\fill}}cccccccc@{}}
            \toprule
            \multirow{2}{*}{\centering Method}
            & \multicolumn{6}{c}{MatrixCity}
            \\
            \cmidrule[0.5pt](lr){2-7}
            {\ }
            & $\text{PSNR}_\uparrow $
            & $\text{SSIM}_\uparrow $
            & $\text{LPIPS}_\downarrow $
            & $\text{\#GS/\#Proxy}_\downarrow $
            & $\text{FPS}_\uparrow $
            & $\text{Time~[h]}_\downarrow $
            \\
            \midrule
            3DGS$^1$ \cite{vanilla_3DGS}
            & \cellcolor{color_2nd}27.03
            & 0.808
            & 0.310
            & 10206~K
            & 66
            & 2.60
            \\
            \midrule
            Scaffold-GS \cite{Scaffold-GS}
            & 26.82
            & 0.811
            & \cellcolor{color_2nd}0.288
            & 4829~K
            & 113
            & \cellcolor{color_2nd}2.31
            \\
            Octree-GS \cite{Octree-GS}
            & 25.18
            & 0.738
            & 0.397
            & \cellcolor{color_1st}4082~K
            & \cellcolor{color_1st}128
            & 2.61
            \\
            Octree-GS* \cite{Octree-GS}
            & 26.61
            & \cellcolor{color_2nd}0.812
            & 0.292
            & 8010~K
            & 79
            & -
            \\
            \midrule
            LC-WSR \cite{Sort-Free}
            & 26.34
            & 0.782
            & 0.340
            & 7841~K
            & 34
            & 3.86
            \\
            \midrule
            Ours $(K=5)$
            & \cellcolor{color_1st}27.40
            & \cellcolor{color_1st}0.826
            & \cellcolor{color_1st}0.260
            & \cellcolor{color_2nd}4165~K
            & \cellcolor{color_2nd}119
            & \cellcolor{color_1st}2.21
            \\
            \bottomrule
        \end{tabular*}
        \begin{tablenotes}
            \footnotesize
            \item[*] Officially released pretrained model.
        \end{tablenotes}
        \end{threeparttable}
    \end{center}
\end{table}

\subsection{Experimental Settings}
\subsubsection{Datasets}
We conducted extensive experiments on scenes from multiple public datasets, adopting the standard configurations established by 3DGS \cite{vanilla_3DGS}.

Our evaluation encompasses 13 standard benchmarks: nine scenes from Mip-NeRF360~\cite{mipnerf360}, two from Tanks\&Temples~\cite{tandt}, and two from DeepBlending~\cite{db}. To further assess robustness, we tested on 10 challenging scenes from BungeeNeRF~\cite{bungeenerf}, featuring multi-scale outdoor environments, and VR-NeRF~\cite{VRNeRF}, containing large-scale intricate interiors. Additionally, we included the large-scale aerial scene Block\_All scene from MatrixCity~\cite{matrixcity}, which comprises 5,621 training and 741 testing images covering an urban area of $2.7~\text{km}^2$.

All experiments were performed at a resolution of 1K.




\subsubsection{Metrics}
We evaluate the reconstruction fidelity with PSNR, SSIM~\cite{SSIM} and LPIPS~\cite{LPIPS}, which are commonly used in image quality measurement.
We also record the model size, frame per second (FPS) and runtime breakdown as indicators of storage overhead and rendering speed tested on RTX 4090 GPU.
Additionally, the consumptions of the radix sort are listed to address the effectiveness of our method in reducing the sorting overhead comparing with $\alpha$-blending, which is a bottleneck for edge devices.
We also report FPS on a laptop with RTX 3060 GPU to verify the applicability of our method in mid-range devices.

\subsubsection{Baselines}
We compare our method with several SOTA approaches, including 3DGS, Mip-NeRF360, Scaffold-GS, Octree-GS, LC-WSR, and StocSplats. Among these, LC-WSR and StocSplats serve as representatives of the OIT based rendering paradigm for Gaussian splatting, while the remaining methods employ the conventional $\alpha$-blending pipeline used in vanilla 3DGS.

The results for StocSplats are sourced directly from its original publication, which reports performance on the Mip-NeRF 360 dataset using an RTX 4090 GPU. As its training code and pre-trained models are not publicly available, we are unable to evaluate its performance on other benchmarks; the corresponding entries in our comparisons are therefore left blank.


\subsubsection{Implementation Details}
\label{sec:implementation_details}
All baseline models are optimized using a combined loss function comprising the $\ell_1$ and SSIM losses. The weighting factor $\lambda_1$ in Eq.~\ref{eq:loss} is set to 0.2. In contrast, a perceptual loss is computed every 20 training iterations with a weight $\lambda_2$ of 0.5.
For neural Gaussian-based methods, including Scaffold-GS, Octree-GS, and our proposed Duplex-GS, experiments are conducted with encoding dimensions $K$ set to 5 and 10, respectively.

Small-scale scenes are trained for 40,000 iterations, with densification concluding at iteration 25,000. Large-scale scenes are trained for 100,000 iterations, with densification ending at iteration 50,000.




\subsection{Results Analysis}
We first validate our method by showing that it achieves competitive accuracy across all datasets while maintaining high efficiency. The results are shown in Tab.~\ref{tab:res1} and Tab.~\ref{tab:res2}.

\begin{figure*} [t]
    \centering
    \includegraphics[width=1.0\linewidth]{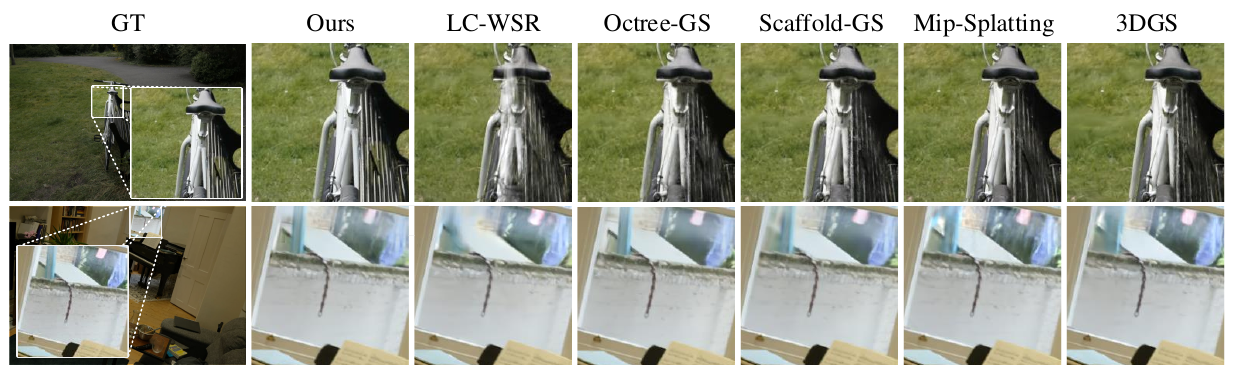}
    \caption{Qualitative comparisons of Duplex-GS and competing baselines on Mip-NeRF360~\cite{mipnerf360} dataset. The proposed Duplex-GS successfully removes the ``transparency" artifacts for the WSR rendering kernel while maintaining the ability to reconstruct fine details.}
    \label{fig:quality_mipnerf}
\end{figure*}

\subsubsection{Quantitative Results}
Tab.~\ref{tab:res1} presents a comprehensive comparison of our method with several SOTA baselines on three real-world datasets. On the Mip-NeRF360 dataset, our approach achieves the highest PSNR when $K=10$, outperforming all baselines. While Mip-Splatting and Octree-GS attain slightly higher SSIM values (0.816), our method yields competitive SSIM (0.813) and achieves improved perceptual quality, as evidenced by a lower LPIPS (0.216). Remarkably, our model also demonstrates strong storage efficiency, requiring significantly less memory than methods such as 3DGS and Mip-Splatting, both of which demand over 750 MB.

On the Tanks\&Temples and DeepBlending datasets, although the PSNR values of our approach are marginally lower, it delivers superior visual quality, attaining the highest SSIM or lowest LPIPS compared to other models. Notably, when compared to LC-WSR, the OIT variant of Gaussian Splatting, our method achieves better performance across all metrics and datasets, showing the effectiveness of the OIT paradigm in Gaussian Splatting on CUDA based high-end GPUs.

To further assess our method under challenging scenarios involving multi-scale training views and large-scale datasets, we conduct experiments on the BungeeNeRF, VR-NeRF and MatrixCity datasets (results shown in Tab.~\ref{tab:res2} and Tab.~\ref{tab:MatrixCity}). On BungeeNeRF, our approach achieves the highest PSNR and lowest LPIPS when $K=10$, and remains highly competitive with $K=5$, attaining the smallest model size among all baselines. The two scenes within VR-NeRF are large-scale, intricate indoor environments containing thousands of training images. Experimental results indicate that OIT based rendering consistently outperforms sort-based $\alpha$-blending approaches for such cases.
On the MatrixCity dataset, our method achieves the highest accuracy across all metrics (PSNR/SSIM/LPIPS) while maintaining competitive efficiency with the most efficient baseline (128~FPS vs. 119~FPS) and the smallest training time. 

\subsubsection{Qualitative Results}
\begin{figure*}
    \centering
    \includegraphics[width=1.0\linewidth]{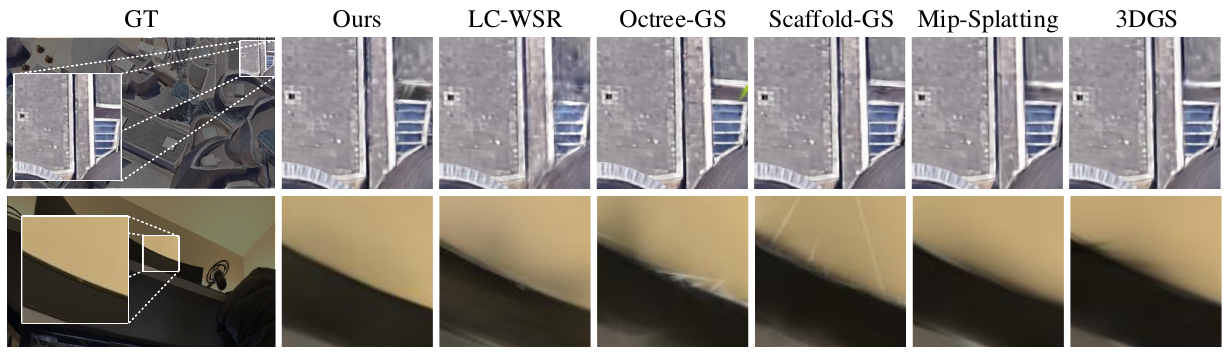}
    \caption{Qualitative comparisons of Duplex-GS and competing baselines on BungeeNeRF~\cite{mipnerf360} and VR-NeRF~\cite{VRNeRF} datasets. the proposed Duplex-GS successfully captures fine details, even in textureless and marginal regions.}
    \label{fig:quality_hard_case}
\end{figure*}

\begin{figure}[t]
    \centering
    \includegraphics[width=1.0\linewidth]{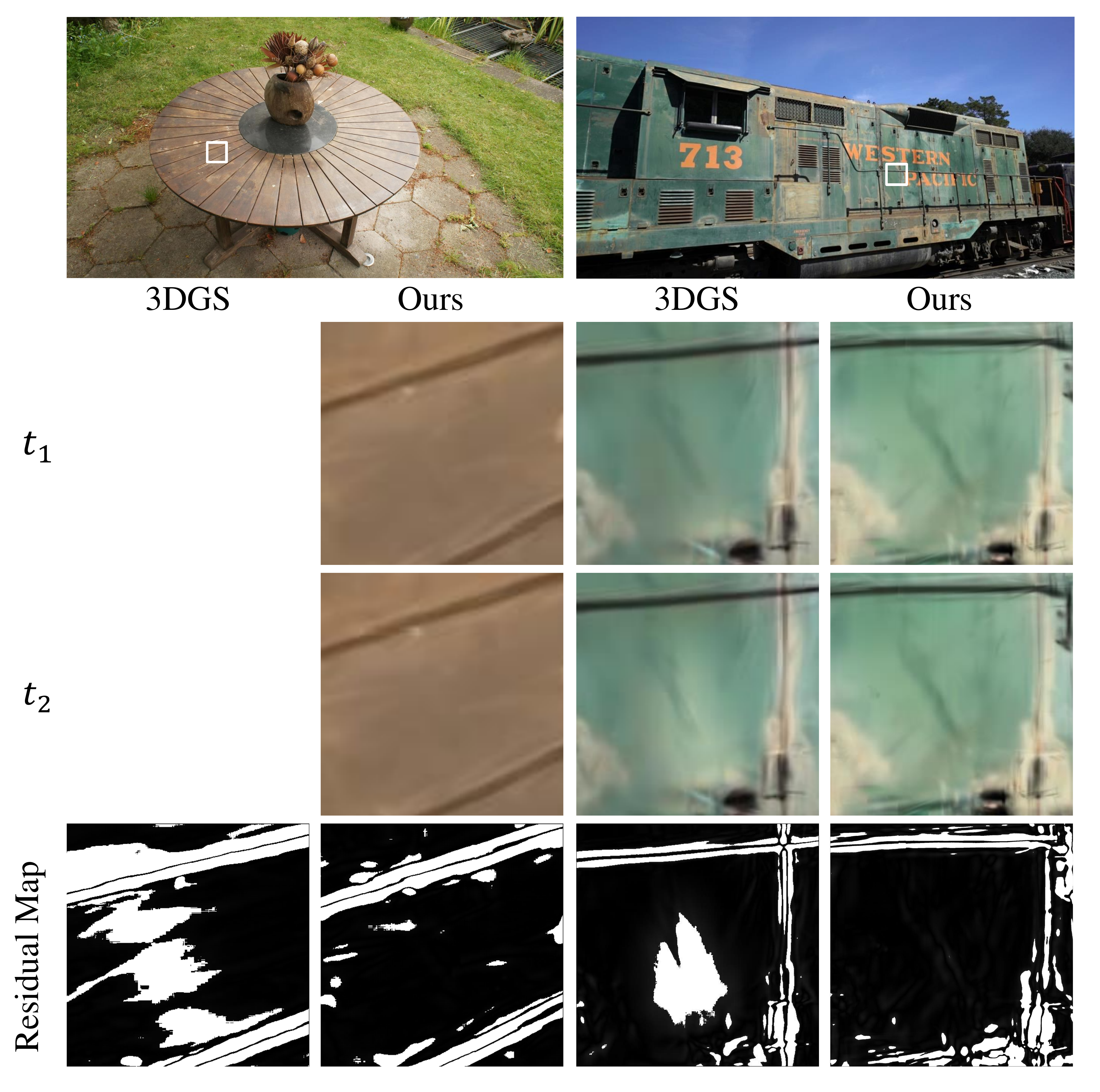}
    \caption{Illustration of ``popping" artifacts removal by recording the differences between contiguous timestep.}
    \label{fig:popping}
\end{figure}

\begin{figure}[t]
    \centering
    \includegraphics[width=1.0\linewidth]{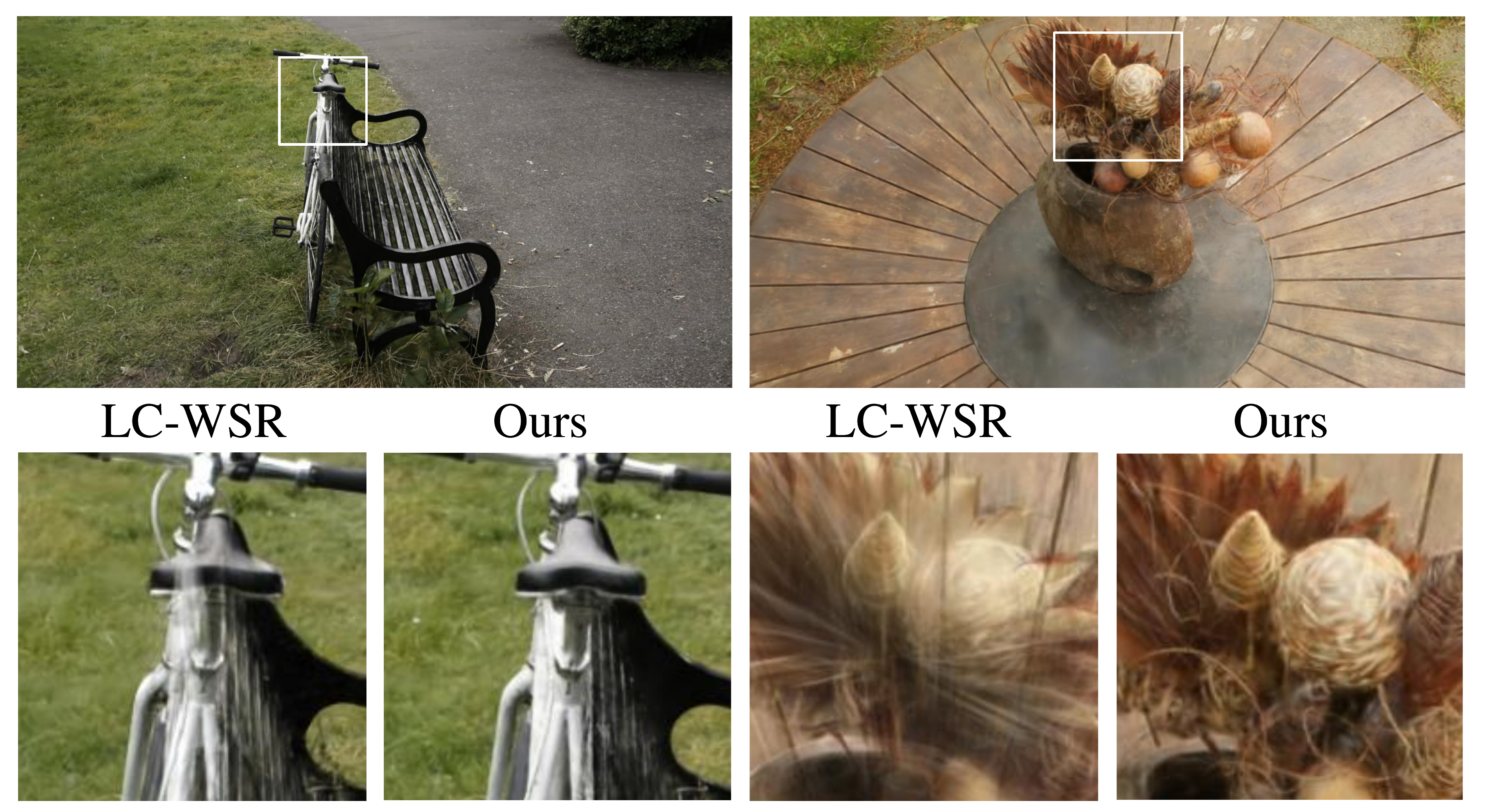}
    \caption{Illustration of ``transparency" artifacts removal.}
    \label{fig:transparency}
\end{figure}

We present qualitative results and comparisons with SOTA baselines to demonstrate the effectiveness of our proposed method in artifact removal and detail preservation. 
As illustrated in Fig.~\ref{fig:quality_mipnerf} and Fig.~\ref{fig:transparency} using examples from the Mip-NeRF360 dataset, our method eliminates the transparency artifacts observed in the previous OIT based LC-WSR, while maintaining intricate scene details. For multi-scale and large-scale scenarios, as shown in Fig.~\ref{fig:quality_hard_case}, the proposed Duplex-GS successfully captures fine details, even in texture-less and marginal regions. Additionally, our approach effectively removes popping artifacts, benefiting from the OIT blending kernel (see Fig.~\ref{fig:popping}). 

\begin{table}[t]
    \setlength{\tabcolsep}{3pt}
    \caption{
    Efficiency analysis of Duplex-GS and competing baselines, reporting the FPS at 1K resolution on RTX 4090 GPU.
    }
    \label{tab:FPS}
    \begin{center}
        \begin{tabular*}{\linewidth}{@{\extracolsep{\fill}}ccccc@{}}
            \toprule
            Method
            & Mip-NeRF360
            & Tan.\&Tem.
            & DeepBlend.
            & BungeeNeRF
            \\
            \midrule
            3DGS \cite{vanilla_3DGS}
            & 149
            & 142
            & 145
            & 52
            \\
            Scaffold-GS \cite{Scaffold-GS}
            & 150
            & 129
            & 185
            & 72
            \\
            Octree-GS \cite{Octree-GS}
            & \cellcolor{color_2nd}183
            & \cellcolor{color_1st}159
            & \cellcolor{color_1st}271
            & \cellcolor{color_2nd}114 
            \\
            \midrule
            LC-WSR \cite{Sort-Free}
            & 77
            & 89
            & 114
            & 31
            \\
            StocSplats$^3$ \cite{kv2025stochasticsplats} 
            & 125
            & -
            & -
            & -
            \\
            \midrule
            Ours $(K=5)$
            & \cellcolor{color_1st}184
            & \cellcolor{color_2nd}147
            & \cellcolor{color_2nd}232
            & \cellcolor{color_1st}124
            \\
            \bottomrule
        \end{tabular*}
    \end{center}
\end{table}

\begin{table*}[tb]
    \setlength{\tabcolsep}{3pt}
    \caption{
    Runtime Breakdown (ms) on MipNeRF360.
    }
    \label{tab:breakdown}
    \begin{center}
        \begin{threeparttable}
        \begin{tabular*}{\linewidth}{@{\extracolsep{\fill}}ccccc cccc ccc@{}}
            \toprule
            $\text{Method}$
            & $\text{Stage}$
            & $\text{bicycle}$
            & $\text{bonsai}$
            & $\text{counter}$
            & $\text{flowers}$
            & $\text{garden}$
            & $\text{kitchen}$
            & $\text{room}$
            & $\text{stump}$
            & $\text{treehill} $
            & $\text{AVE} $
            \\
            \midrule
            \multirow{3}{*}{\centering LC-WSR}
            & rasterization
            & 7.0177
            & 2.7192
            & 3.2817
            & 3.8094
            & 7.0134
            & 4.5389
            & 2.8045
            & 4.1043
            & 5.7666
            & 4.5618
            \\
            {\ }
            & blending
            & 10.5629
            & 4.2739
            & 4.0905
            & 4.9504
            & 8.9972
            & 6.0351
            & 4.7305
            & 5.2737
            & 8.6236
            & 6.5042
            \\
            {\ }
            & sum
            & 17.5806
            & 6.9932
            & 8.3722
            & 8.7599
            & 16.0106
            & 10.5741
            & 7.5350
            & 9.3779
            & 14.3903
            & 11.0660
            \\
            \midrule
            \multirow{5}{*}{\centering Octree-GS}
            & decode
            & 0.0027
            & 0.0024
            & 0.0030
            & 0.0035
            & 0.0038
            & 0.0035
            & 0.0018
            & 0.0019
            & 0.0021
            & 0.0027
            \\
            {\ }
            & rasterization
            & 0.4232
            & 0.3942
            & 0.4723
            & 0.3498
            & 0.4280
            & 0.4598
            & 0.4497
            & 0.3083
            & 0.3823
            & 0.4075
            \\
            {\ }
            & sorting
            & 0.6793
            & 0.5985
            & 0.8814
            & 0.4757
            & 0.6933
            & 0.8888
            & 0.7727
            & 0.3011
            & 0.4941
            & 0.6427
            \\
            {\ }
            & blending
            & 1.3343
            & 1.2240
            & 1.4663
            & 1.0070
            & 1.4140
            & 1.5561
            & 1.5236
            & 0.8118
            & 1.0878
            & \cellcolor{color_1st}1.2695
            \\
            {\ }
            & sum
            & 2.4394
            & 2.2191
            & 2.8231
            & 1.8367
            & 2.5391
            & 2.9082
            & 2.7477
            & 1.4230
            & 1.9663
            & 2.3225
            \\
            \midrule
            \multirow{5}{*}{\centering Ours}
            & decode
            & 0.0028
            & 0.0027
            & 0.0028
            & 0.0025
            & 0.0032
            & 0.0030
            & 0.0023
            & 0.0023
            & 0.0023
            & 0.0027
            \\
            {\ }
            & rasterization
            & 0.2562
            & 0.2214
            & 0.2387
            & 0.2047
            & 0.2200
            & 0.2377
            & 0.1944
            & 0.2284
            & 0.2580
            & \cellcolor{color_1st}0.2289
            \\
            {\ }
            & sorting
            & 0.3799
            & 0.4169
            & 0.6360
            & 0.1576
            & 0.2714
            & 0.5857
            & 0.3576
            & 0.1465
            & 0.1925
            & \cellcolor{color_1st}0.3493
            \\
            {\ }
            & blending
            & 1.7981
            & 1.3834
            & 1.8776
            & 0.9951
            & 1.6763
            & 1.6698
            & 2.1289
            & 0.8722
            & 1.3549
            & 1.5285
            \\
            {\ }
            & sum
            & 2.4371
            & 2.0243
            & 2.7552
            & 1.3600
            & 2.1709
            & 2.4963
            & 2.6832
            & 1.2493
            & 1.8078
            & \cellcolor{color_1st}2.1093
            \\
            \bottomrule
        \end{tabular*}
        \end{threeparttable}
    \end{center}
\end{table*}

\subsubsection{Efficiency Analysis}
Tab.~\ref{tab:FPS} provides a comparison of rendering speeds, measured in FPS, between our method and several state-of-the-art baselines across four benchmark datasets. Our approach consistently achieves the highest or highly competitive FPS on all datasets. On the Mip-NeRF360 dataset, our method achieves 184 FPS, surpassing all other baselines and comparable to the fastest prior approach, Octree-GS (183 FPS). For the Tanks\&Temples dataset, our method obtains 147 FPS, demonstrating performance close to the best-performing baseline. While on the DeepBlending dataset, our FPS of 232 is similarly on par with the leading method. Most notably, on the BungeeNeRF dataset, our method achieves 124 FPS, a considerable improvement over existing baselines. 
These results highlight the efficiency and scalability of our approach, demonstrating real-time rendering capability and practical superiority in intricate and multi-scale scenarios.

Tab.~\ref{tab:breakdown} presents a runtime breakdown to elucidate the mechanisms of our hybrid renderer. The analysis reveals that while LC-WSR benefits from hardware graphics pipeline compatibility, it performs suboptimally on TBR architectures due to redundant Gaussian primitives introduced in rendering (exemplified in Fig.~\ref{fig:T}) and a lack of physical constraints.
In contrast, our method achieves a significant acceleration in the rasterization and sorting stages. This gain is directly attributable to our coarse-grained processing of cell proxies, which drastically reduces the number of primitives handled compared to per-Gaussian methods.
This efficiency introduces a trade-off: the blending phase exhibits increased latency. This is an expected consequence of our design, which retains sequential blending for cells and may process redundant Gaussians by operating on cell-based groups (of size $K$) rather than individual Gaussians. Despite this localized overhead, our approach achieves the highest overall rendering speed, demonstrating the effectiveness of the proposed hybrid paradigm.


Radix sort is employed in 3DGS to determine the explicit ordering of Gaussian primitives, sorting $b$-digit numbers with $m$ possible values in $\mathcal{O}\left(b\cdot(n+m)\right)$ time and $\mathcal{O}(n+m)$ space. Notably, the sequence length $n$ contributes linearly to the overall complexity. To evaluate the effectiveness of our approach in reducing the computational and memory requirements of view-adaptive sorting, we record the sorting length and corresponding memory consumption during the radix sort procedure. Compared with methods adhering to conventional $\alpha$-blending paradigm, the results, presented in Tab.~\ref{tab:radix_sort}, show that our method successfully mitigates the sorting overhead by introducing cell proxies, thereby demonstrating superior efficiency.

\subsubsection{Hyperparameter Analysis}
\begin{figure}
    \centering
    \includegraphics[width=1.0\linewidth]{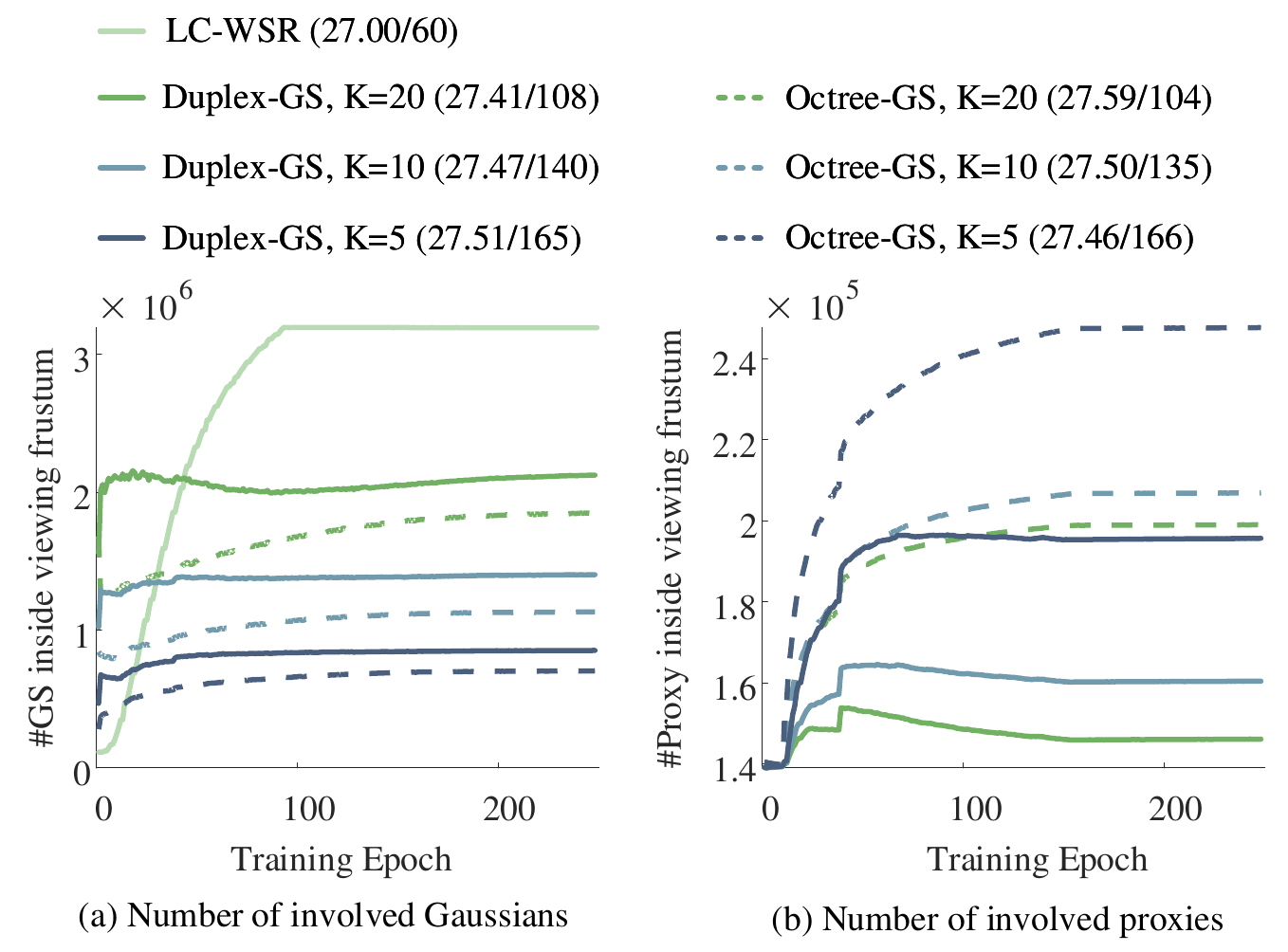}
    \caption{Performance analysis with different hyperparameter $K$ on garden scenario from Mip-NeRF 360 dataset, which calculates metric (PSNR~[dB]/FPS), the number of valid Gaussians and proxies inside a same viewing frustum during training.
    (a) The valid Gaussians is number of Gaussians with $\sigma>0$, indicating the efficiency of blending phase.
    (b) The involved proxies is the number of cells or anchors located inside the viewing frustum. The proposed cell is more sparse than anchors, resulting in acceleration in rasterization stage.
    }
    \label{fig:K}
\end{figure}

We give analysis on the influence on the hyperparameter $K$, which controls the number of Gaussians decoded per cell.
Since the opacity $\sigma$, decoded via $F_\sigma$ and activated by a Tanh function, must satisfy $\sigma>0$ for a Gaussian to be considered valid and included in rendering, it directly regulates the number of active primitives.
The results show that while varying $K$ leads to only minor deviations in reconstruction accuracy, it substantially impacts computational efficiency. As illustrated in Fig.~\ref{fig:K}, increasing $K$ results in more Gaussians being activated during rendering, which slows down the blending stage (Fig.~\ref{fig:K}-a). Conversely, a higher $K$ reduces the number of proxies involved, thereby accelerating the rasterization process (Fig.~\ref{fig:K}-b).
This trade-off elucidates the performance characteristics of our hybrid renderer: the slower blending speed compared to $\alpha$-blending-based frameworks and significant acceleration over LC-WSR are attributable to the number of valid Gaussians (Fig.~\ref{fig:K}-a), 
Furthermore, the results confirm that our cell-based framework exploits spatial sparsity more effectively than anchor-based approaches, as evidenced by the fewer proxies located within the same viewing frustum (Fig.~\ref{fig:K}-b).

\begin{table}[t]
    \setlength{\tabcolsep}{3pt}
    \caption{
    Comparisons about the overhead of radix sort stage. The proposed cell rasterization consistently reduce both the time and space consumptions compared with $\alpha$-blending based methods.
    }
    \label{tab:radix_sort}
    \begin{center}
        \begin{tabular*}{\linewidth}{@{\extracolsep{\fill}}ccccc@{}}
            \toprule
            \multirow{2}{*}{\centering Method}
            & \multicolumn{2}{c}{Mip-NeRF360}
            & \multicolumn{2}{c}{BungeeNeRF}
            \\
            \cmidrule[0.5pt](lr){2-3} \cmidrule[0.5pt](lr){4-5}
            {\ }
            & $\text{Sorting Length}_\downarrow $
            & $\text{Memory}_\downarrow $
            & $\text{Sorting Length}_\downarrow $
            & $\text{Memory}_\downarrow $
            \\
            \midrule
            3DGS \cite{vanilla_3DGS}
            & 7,000,602
            & 81.0\,MB
            & 21,798,876
            & 255.0\,MB
            \\
            Scaffold-GS \cite{Scaffold-GS}
            & 4,588,785
            & 53.69\,MB
            & 6,901,347
            & 80.74\,MB
            \\
            Octree-GS \cite{Octree-GS}
            & 4,701,055
            & 55.09\,MB
            & 5,480,410 
            & 64.14\,MB
            \\
            Ours
            & \cellcolor{color_1st}3,217,845
            & \cellcolor{color_1st}37.7\,MB
            & \cellcolor{color_1st}2,843,604
            & \cellcolor{color_1st}33.3\,MB
            \\
            \bottomrule
        \end{tabular*}
    \end{center}
\end{table}

\subsection{Ablation Studies}
\begin{figure*}
    \centering
    \includegraphics[width=1.0\linewidth]{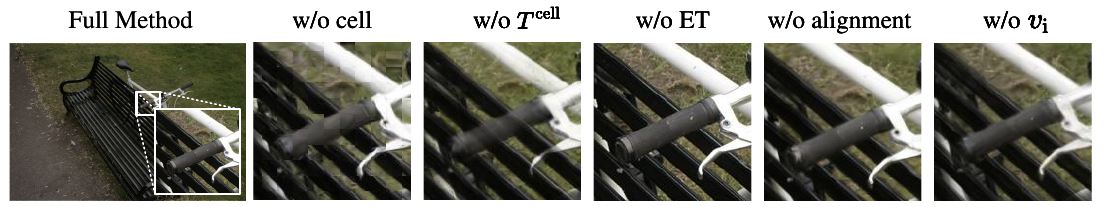}
    \caption{Visualization of ablation study. Quantitative results are shown in Tab.~\ref{tab:ablation}.}
    \label{fig:ablation}
\end{figure*}
\subsubsection{Cell vs. Anchor}
The introduction of the ellipsoidal cell represents a key innovation, enabling efficient proxy rasterization and physical grounded WSR-based blending. In contrast, anchors adhere to the conventional $\alpha$-blending pipeline, which processes Gaussians individually. To validate the effectiveness of cells in proxy rasterization, we conduct an ablation study by replacing cells with anchor points while maintaining our hybrid rendering paradigm and retraining the models. As shown in Tab.~\ref{tab:ablation}, 
this anchor-based rasterization fails due to the lack of well-defined visible regions, confirming the necessity of our geometric cell structure.

\subsubsection{Cell-level Transmittance}
The physically grounded transmittance, shown in Eq.~\ref{eq:dulpex_T}, is the key innovation which eliminates transparency artifacts as well as enable early termination. We replace it with the conventional LC-WSR blending kernel in the proposed pipeline, which is marked as ``w/o $T^\text{cell}$" in Tab.~\ref{tab:ablation} and Fig.~\ref{fig:ablation}.
The absence of physical constraints result in transparency artifacts and degraded accuracy.

\subsubsection{Early Termination (ET)}
\begin{figure}[t]
    \centering
    \includegraphics[width=1.0\linewidth]{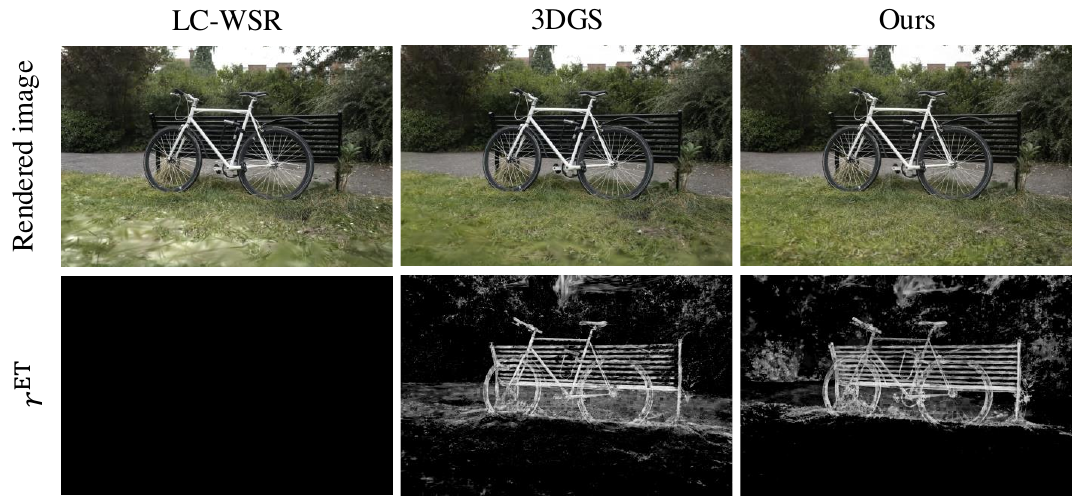}
    \caption{Visualization of the early termination mechanism.
    The existing OIT 3DGS method, LC-WSR, does not support early termination due to its sort-free design. In contrast, Duplex-GS achieves termination results comparable to those of vanilla 3DGS, without requiring an explicit Gaussian order.}
    \label{fig:ET}
\end{figure}

The introduction of the early termination mechanism is a key advantage of our method over LC-WSR, which 
selectively omit computations that have negligible impact on the final rendered output with extremely small transmittance.
To visualize the mechanism, we record the termination ratio with
\begin{equation}
    r^\text{ET} = 1 - \frac{N^\text{rendered}}{N^\text{val}},
\end{equation}
where $N^\text{rendered}$ denotes the number of Gaussians contributed to the final color, $N^\text{val}$ denotes that of the total interacted valid Gaussians along the ray. 
Since the proposed transmittance achieves a decay pattern similar to $\alpha$-blending (Fig.~\ref{fig:T}), we maintain the same threshold value as used in vanilla 3DGS.
The results are visualized in Fig.~\ref{fig:ET}.
Our method demonstrates termination behavior which closely matches that of $\alpha$-blending, whereas prior sort-free Gaussian Splatting methods lack this feature. To evaluate its effectiveness, we conduct an ablation study by disabling early termination during model training. As shown in Tab.~\ref{tab:ablation}, enabling early termination leads to improvements in both accuracy and efficiency.

\begin{table}[tb]
    \setlength{\tabcolsep}{3pt}
    \caption{
    Ablation experiments. Visual examples are shown in Fig.~\ref{fig:ablation}.
    }
    \label{tab:ablation}
    \begin{center}
        \begin{tabular*}{\linewidth}{@{\extracolsep{\fill}}ccccc@{}}
            \toprule
            {\centering Method}
            & $\text{PSNR}_\uparrow $
            & $\text{SSIM}_\uparrow $
            & $\text{LPIPS}_\downarrow $
            & $\text{FPS}_\uparrow $
            \\
            \midrule
            w/o cell
            & 19.30
            & 0.655
            & 0.365
            & \cellcolor{color_1st}270
            \\
            w/o $T^\text{cell}$
            & 26.60
            & 0.790
            & 0.231
            & 212
            \\
            w/o ET
            & 27.66
            & 0.801
            & 0.220
            & 155
            \\
            w/o alignment
            & 27.54
            & 0.799
            & 0.223
            & 229
            \\
            w/o $v_n$
            & 27.59
            & 0.801
            & \cellcolor{color_1st}0.218
            & 174
            \\
            \midrule
            Full method
            & \cellcolor{color_1st}27.74
            & \cellcolor{color_1st}0.802
            &\cellcolor{color_1st} 0.218
            & 184
            \\
            \bottomrule
        \end{tabular*}
    \end{center}
\end{table}

\subsubsection{Alignment Restriction}
Unlike Gaussians whose geometry is directly optimized through blending operations, cells are updated guided by Gaussians and several constraints (Eq.~\ref{eq:quaternion} and Eq.~\ref{eq:scale_res}) are imposed to align coverage between cell and the corresponding Gaussians.
To show the effectiveness of these conditions, we mute these restriction and the results are shown in Tab.~\ref{tab:ablation}.
The accuracy decreases as expected, however, the rendering speed is improved, which offers a solution for resource-constrained platform for its significant acceleration while maintaining competitive visual quality. 

\subsubsection{Order Confusion}
As the sorting results of cell level cannot reflect the exact order of Gaussians. we introduce learnable parameter $v_n$ to compensate the ambiguity. We train models without $v_n$ to show its effectiveness. The results are shown in Tab.~\ref{tab:ablation}. By introducing $v_n$, the accuracy is improved as more degree-of-freedom is introduced to handle the overlap ambiguity among cells. 

\begin{table}[t]
    \setlength{\tabcolsep}{3pt}
    \caption{
    Efficiency analysis of Duplex-GS and competing baselines on Mip-NeRF360 datasets, reporting FPS at 1K resolution on a laptop with RTX 3060 GPU .
    }
    \label{tab:FPS_3060}
    \begin{center}
        \begin{tabular*}{\linewidth}{@{\extracolsep{\fill}}cccccc@{}}
            \toprule
            Method
            & 3DGS
            & LC-WSR
            & Octree-GS
            & Ours ($K$=5)
            & Ours (w/o alignment)
            \\
            \midrule
            FPS
            & 42
            & 21
            & 47
            & 49
            & 72
            \\
            \bottomrule
        \end{tabular*}
    \end{center}
\end{table}
\subsection{Performance on Mid-range GPUs}
We test rendering performance on a laptop with RTX 3060 GPU to validate the practicability of our method on mid-range devices. The results are shown in Tab.~\ref{tab:FPS}. Our method consistently achieves highest rendering speed while maintaining competitive visual quality.

\section{Limitation and Conclusion}
This work presents a dual-hierarchy Gaussian model that effectively resolves popping artifacts inherent in vanilla $\alpha$-blending and the transparency artifacts of WSR. By introducing geometric proxy and a hybrid rendering paradigm, our approach substantially reduces the overhead associated with global sorting while achieving real-time performance as well as competitive visual quality.

Several limitations suggest directions for future work. First, the geometric relationships between cells and their constituent Gaussians require further optimization, as imperfect spatial alignment can lead to performance degradation and redundant calculation, as illustrated in Fig.~\ref{fig:false_positive_skip}. 
Second, the core WSR operation within each cell is inherently order-independent, presenting substantial, yet unexplored, parallel computing opportunities. Custom hardware accelerators based on FPGA or ASIC can capitalize on this parallelism to achieve higher performance and energy efficiency compared to general-purpose GPUs, thereby facilitating deployment on resource-constrained edge devices.

\bibliographystyle{IEEEtran}
\bibliography{IEEEabrv,transbib}


 




\vfill

\vspace{30pt}

\begin{IEEEbiography}[{\includegraphics[width=1in,height=1.25in,clip,keepaspectratio]{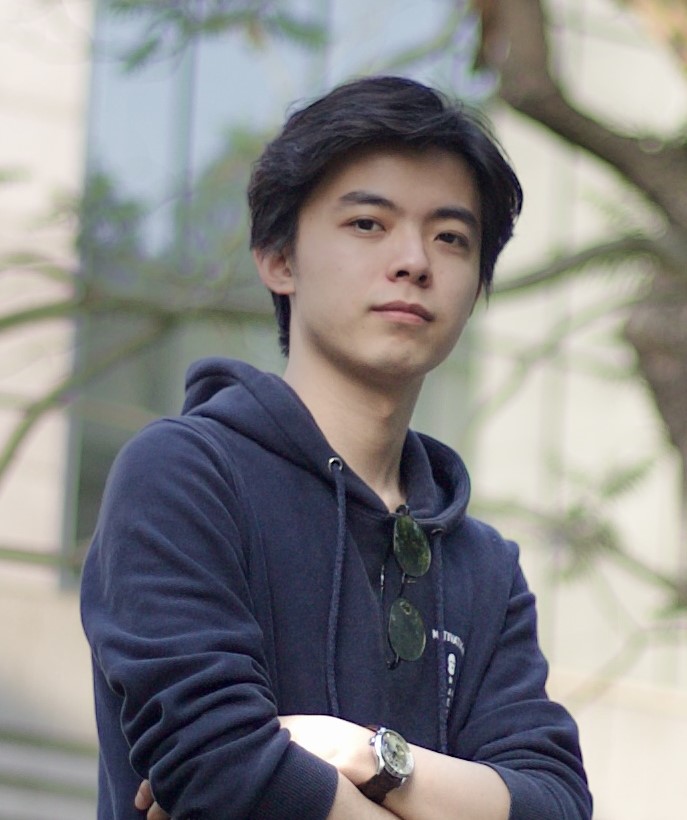}}]{Weihang Liu}
received the B.S. and M.E. degree in electronics and communication engineering in 2019 and 2022. He is currently working toward the doctoral degree in electronic and information engineering with School of Information Science and Technology, ShanghaiTech University, Shanghai, China. His research interests include computer graphics, computer vision and artificial intelligence.
\end{IEEEbiography}

\vspace{-10pt}

\begin{IEEEbiography}[{\includegraphics[width=1in,height=1.25in,clip,keepaspectratio]{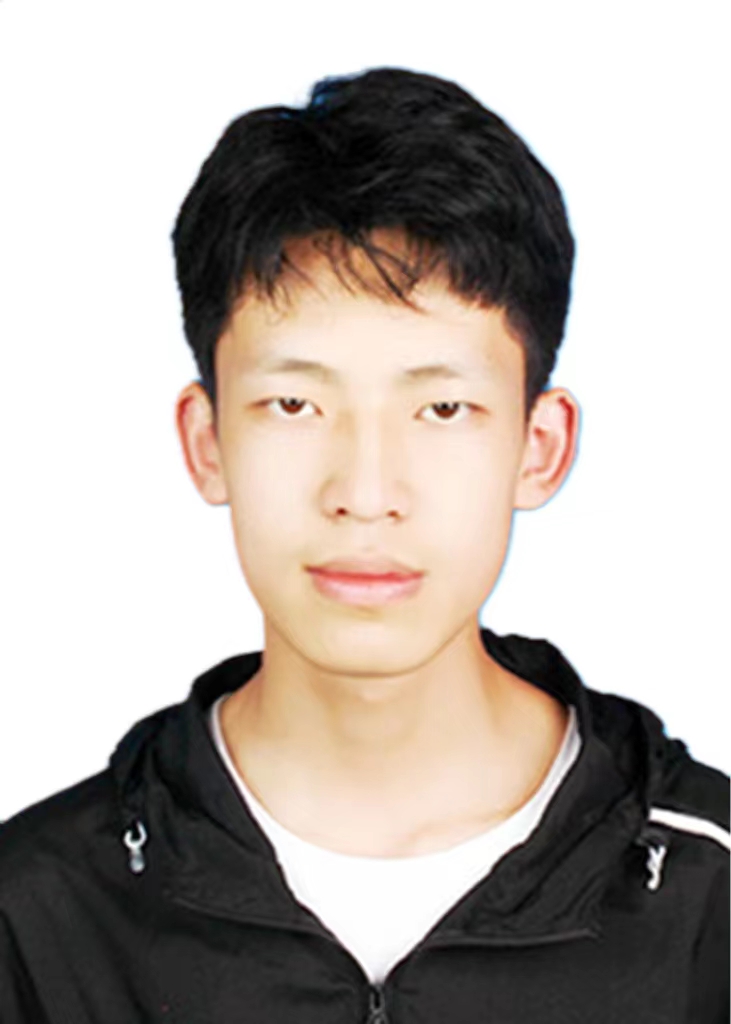}}]{Yuke Li}
received the B.S. degree in data science from China University of Petroleum, Qingdao, China, in 2024. He is currently pursuing the master's degree with ShanghaiTech University, Shanghai, China.
His research interests include AI, neural networks, and data science.
\end{IEEEbiography}

\vspace{-10pt}

\begin{IEEEbiography}[{\includegraphics[width=1in,height=1.25in,clip,keepaspectratio]{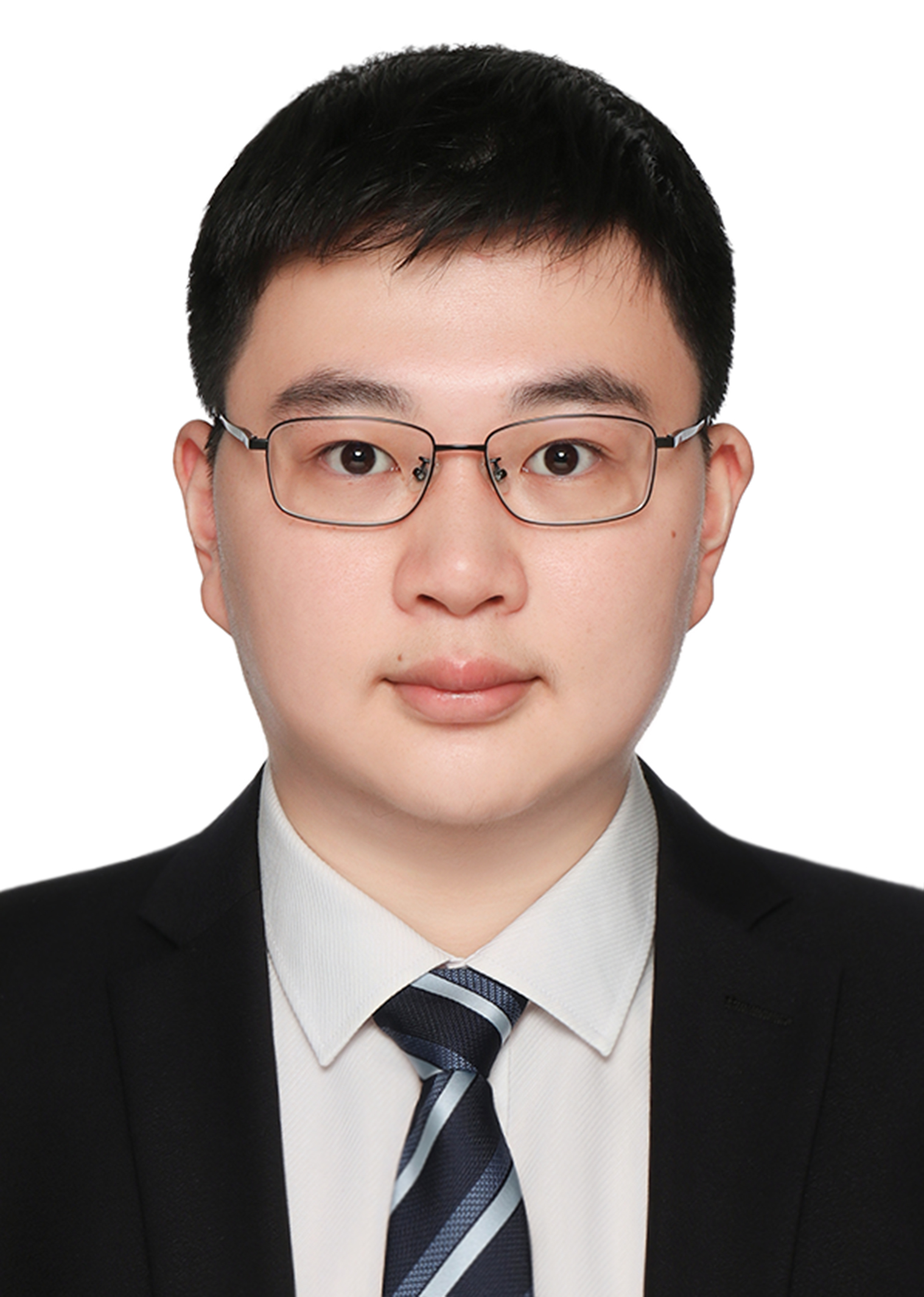}}]{Yuxuan Li}
received the B.S. degree in microelectronics science and engineering from Shanghai University in 2025. He is currently pursuing the doctoral degree at ShanghaiTech University. His research interests include computer graphics and computer architecture.
\end{IEEEbiography}

\vspace{-10pt}

\begin{IEEEbiography}[{\includegraphics[width=1in,height=1.25in,clip,keepaspectratio]{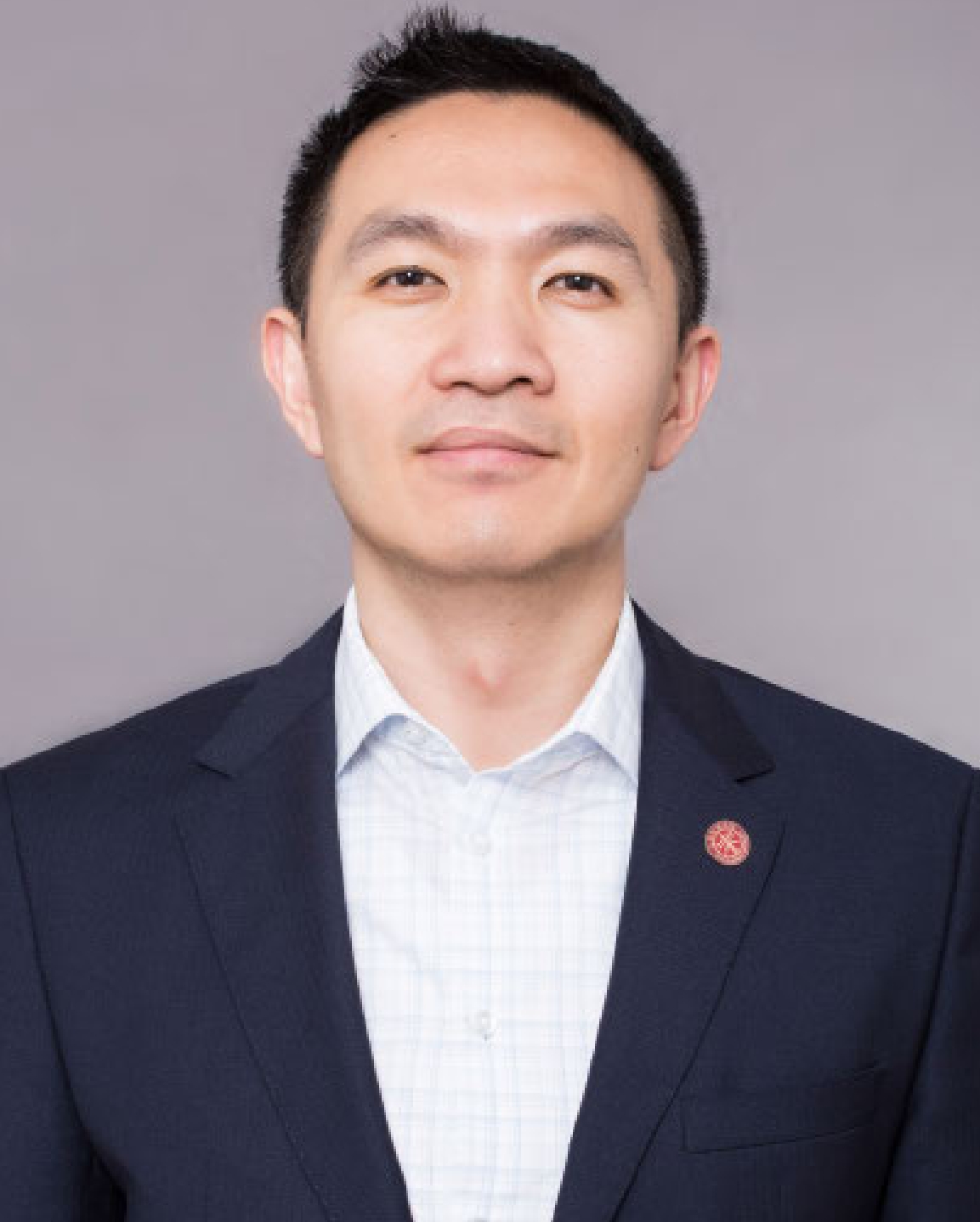}}]{Jingyi Yu}
(Fellow, IEEE) received the B.S. degree from Caltech in 2000 and the Ph.D. degree from MIT in 2005. He is currently the Vice Provost with ShanghaiTech University. Before joining ShanghaiTech University, he was a Full Professor with the Department of Computer and Information Sciences, University of Delaware. His current research interests include computer vision and computer graphics, especially computational photography and nonconventional optics and camera designs. He is a recipient of the NSF CAREER Award and the AFOSR YIP Award. He served as an Area Chair for many international conferences, including CVPR, ICCV, ECCV, IJCAI, and NeurIPS. He was the Program Chair of CVPR 2021 and will be the Program Chair of ICCV 2025.
\end{IEEEbiography}

\vspace{-10pt}

\begin{IEEEbiography}[{\includegraphics[width=1in,height=1.25in,clip,keepaspectratio]{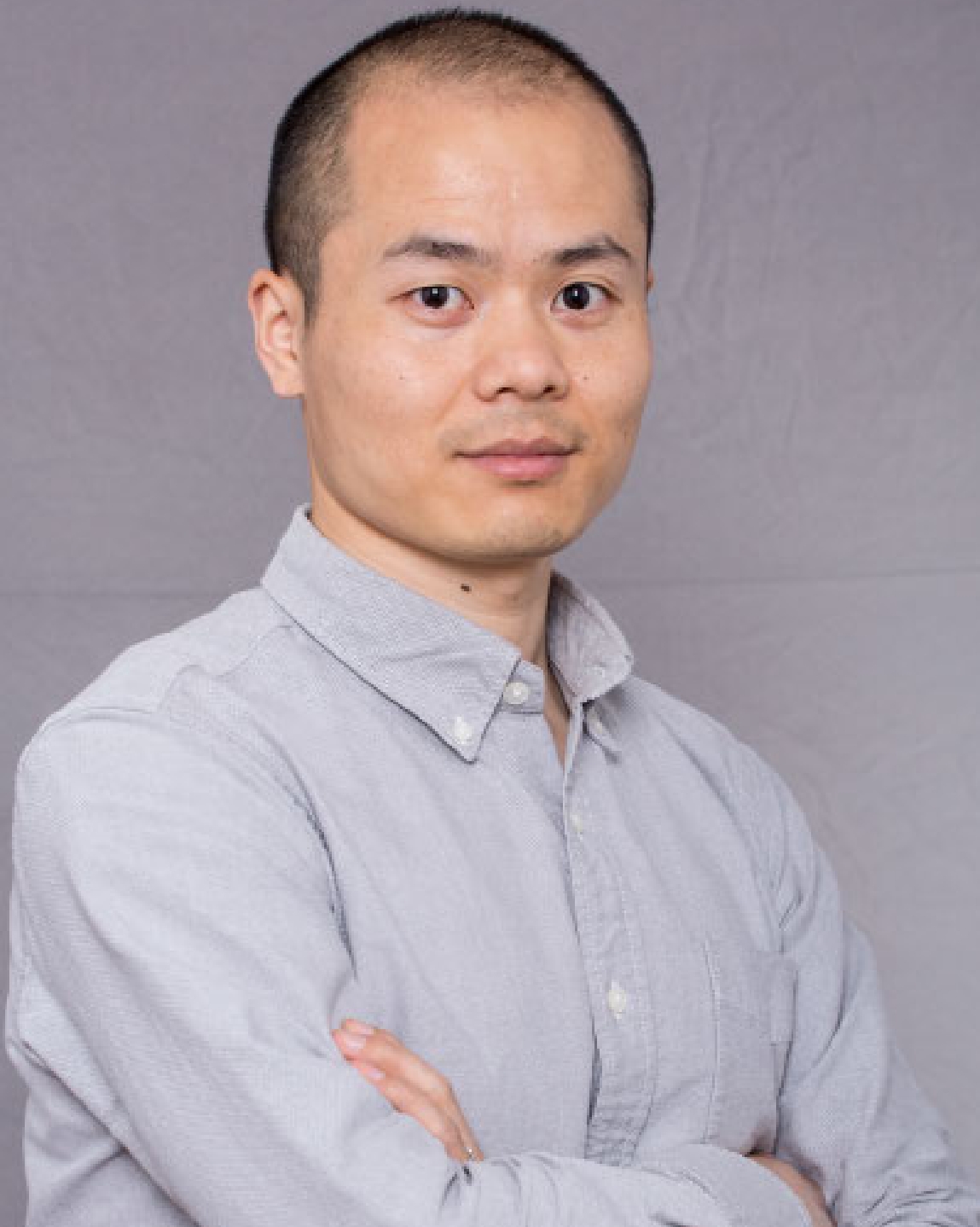}}]{Xin Lou}(Senior Member, IEEE) received the B.Eng. degree in Electronic Information Technology and Instrumentation from Zhejiang University (ZJU), China, in 2010 and M.Sc. degree in System-on-Chip Design from Royal Institute of Technology (KTH), Sweden, in 2012 and PhD degree in Electrical and Electronic Engineering from Nanyang Technological University (NTU), Singapore, in 2016. Then he joined VIRTUS, IC Design Centre of Excellence at NTU as a research scientist. 
He is currently an Associate Professor with the School of Information Science and Technology, ShanghaiTech University, Shanghai, China. His research interests primarily focus on high-performance and energy-efficient integrated circuits and systems for vision and graphics processing. Dr. Lou serves as an Associate Editor of IEEE Transactions on Very Large Scale Integration, and was an Associate Editor of IEEE Transactions on Circuits and Systems II: Express Briefs (2022-2023), a Guest Editor of Associate Editor of IEEE Transactions on Circuits and Systems I (2024).
\end{IEEEbiography}

\end{document}